\documentclass{article}
\usepackage{fancyhdr}
\usepackage{report}

\usepackage{soul}
\usepackage[utf8]{inputenc} % allow utf-8 input
\usepackage[T1]{fontenc}    % use 8-bit T1 fonts
\usepackage{hyperref}       % hyperlinks
\usepackage{url}            % simple URL typesetting
\usepackage{booktabs}       % professional-quality tables
\usepackage{amsfonts}       % blackboard math symbols
\usepackage{fancyhdr}       % custom headers and footers

\usepackage{nicefrac}       % compact symbols for 1/2, etc.
\usepackage{microtype}      % microtypography
\usepackage{graphicx}
\usepackage{pifont}
\usepackage{multirow}
\usepackage{CJKutf8}
\definecolor{darkmagenta}{rgb}{0.56, 0.0, 1.0}
\definecolor{softyellow}{rgb}{1.0, 0.92, 0.3} % richer, warmer yellow
\definecolor{LightAquamarine}{rgb}{0.75, 1.0, 0.8} % soft aqua green
\definecolor{FireBrick}{RGB}{178,34,34}
\definecolor{MediumPurple}{RGB}{147,112,219}

\definecolor{uclablue}{rgb}{0.15, 0.45, 0.68}
\hypersetup{
    breaklinks,
    colorlinks=true,
    citecolor={darkmagenta},
    linkcolor={uclablue},
    urlcolor={uclablue}
}
\usepackage{wrapfig}
\usepackage{float}
\usepackage{subcaption}
\usepackage{placeins}
\usepackage{lipsum} % 用于生成示例文字
\usepackage{tcolorbox}
\usepackage{amsmath}
\usepackage{amssymb}
\usepackage{utfsym}
\usepackage{fontawesome}
\usepackage{xspace}
\usepackage{enumitem}
\usepackage{multirow} 
\tcbuselibrary{breakable}
\usepackage{enumitem}
\usepackage{colortbl}
\usepackage{fancyhdr}

\usepackage{longtable}
\usepackage{tcolorbox}
\tcbuselibrary{skins, breakable}
\usepackage{minted}       % 可选：用于代码高亮（需 Python 和 pygmentize）
\usepackage{xcolor}
\definecolor{mylightblue}{HTML}{FAFAFF}  % 注意：十六进制字母建议大写
\newtcolorbox{examplebox}[1][]{%
    title=#1,
    colback=mylightblue, % 使用浅蓝色的 HTML 颜色代码
    colframe=blue!70!black,
    fonttitle=\bfseries\large,
    top=10pt,
    bottom=10pt,
    left=10pt,
    right=10pt,
    boxsep=5pt,
    boxrule=1pt,
    enhanced,
    breakable,
    before skip=10pt,
    after skip=10pt,
    overlay unbroken and first={%
        \draw[blue!70!black, line width=1pt] (frame.north west) rectangle (frame.south east);
    },
    overlay unbroken and last={%
        \draw[blue!70!black, line width=1pt] (frame.north west) rectangle (frame.south east);
    }
}

\usepackage{tcolorbox}
\usepackage{transparent}
\usepackage[table]{xcolor}
\usepackage{array}    % for custom column types
\usepackage{makecell} % for multi-line cells
\usepackage{adjustbox} % for resizing table
\usepackage{pifont}   % for \cmark and \xmark
\usepackage{multirow}
\newcommand{\cmark}{\textcolor{green!60!black}{\ding{51}}}
\newcommand{\xmark}{\textcolor{red!70!black}{\ding{55}}}

\definecolor{njuPurple}{RGB}{220,205,230}     % 南大紫（深）
\definecolor{njuPurpleLight}{RGB}{250,245,252}   % 极浅的紫色背景（接近白）

\newtcolorbox{abstractbox}{
    colback=njuPurpleLight,   % 浅紫色背景
    colframe=njuPurple,       % 深紫色边框
    boxrule=1pt,              % 边框粗细
    arc=4mm,                  % 圆角
    left=8pt,                 % 左边距
    right=8pt,                % 右边距
    top=8pt,                  % 上边距
    bottom=8pt,               % 下边距
    opacityback=0.95
}

\title{DR$^{3}$-Eval: Towards Realistic and Reproducible Deep Research Evaluation}
\author{
\textbf{Qianqian Xie$^{1}$}\thanks{Equal contribution.},
\textbf{Qingheng Xiong$^{1}$}\footnotemark[1],
\textbf{He Zhu$^{2}$}\footnotemark[1], 
\textbf{Tiantian Xia$^{5}$},
\textbf{Xueming Han$^{3}$},\\
\textbf{Fanyu Meng$^{3}$},
\textbf{Jiakai Wang$^{2}$},
\textbf{Zhiqi Bai$^{2}$},
\textbf{Chengkang Jiang$^{1}$},
\textbf{Zhaohui Wang$^{1}$},\\
\textbf{Yubin Guo$^{1}$},
\textbf{Yuqing Wen$^{4}$},
\textbf{Jiayang Mao$^{1}$},
\textbf{Zijie Zhang$^{1}$},
\textbf{Shihao Li$^{1}$},\\
\textbf{Yanghai Wang$^{1}$},
\textbf{Yuxiang Ren$^{1}$},
\textbf{Junlan Feng$^{3}$}\thanks{Corresponding author},
\textbf{Jiaheng Liu$^{1}$}\thanks{Corresponding author.} \\
\begin{tabular}{c}
{\normalsize $^1$ Nanjing University \quad
$^2$ M-A-P \quad
$^3$ Jiutian Research} \\
{\normalsize $^4$ National University of Singapore
$^5$ Nanjing University of Science and Technology}
\end{tabular} \\
\begin{tabular}{c @{\hspace{2cm}} c}
\texttt{xieqianqian@smail.nju.edu.cn} &
\texttt{liujiaheng@nju.edu.cn}
\end{tabular}
}

\begin{document}

\maketitle

\begin{abstractbox}
\begin{center}
\textbf{\Large Abstract}
\end{center}
Deep Research Agents (DRAs) aim to solve complex, long-horizon research tasks involving planning, retrieval, multimodal understanding, and report generation, yet their evaluation remains challenging due to dynamic web environments and ambiguous task definitions. We propose \textbf{DR$^{3}$-Eval}, a realistic and reproducible benchmark for evaluating deep research agents on multimodal, multi-file report generation. DR$^{3}$-Eval is constructed from authentic user-provided materials and paired with a per-task static research sandbox corpus that simulates open-web complexity while remaining fully verifiable, containing supportive documents, distractors, and noise. Moreover, we introduce a multi-dimensional evaluation framework measuring Information Recall, Factual Accuracy, Citation Coverage, Instruction Following, and Depth Quality, and validate its alignment with human judgments. Experiments with our developed multi-agent system DR$^{3}$-Agent based on multiple state-of-the-art language models demonstrate that DR$^{3}$-Eval is highly challenging and reveals critical failure modes in retrieval robustness and hallucination control. 
Our code and data are publicly available.\footnote{\url{https://github.com/NJU-LINK/DR3-Eval}}
\end{abstractbox}

\section{Introduction}
Recent advances in large language models have enabled the development of Deep Research Agents (DRAs), which aim to autonomously perform complex, long-horizon research tasks involving planning, iterative information retrieval, multimodal understanding, and synthesis of structured, citation-grounded reports~\citep{webthinkerempoweringlargereasoning, multimodaldeepresearchergeneratingtextchart, agentlaboratory,  xu2025comprehensivesurveydeepresearch,openai2025o3, google2025gemini, qwen2025deepresearch, perplexity2024research, grok1_2024, tongyi_deepresearch_2025, deerflow_2025, camel_workforce_2026}. Unlike traditional question-answering systems, DRAs must operate under uncertainty, reason over heterogeneous and noisy information sources, and integrate evidence into coherent analytical outputs. As these capabilities rapidly improve, establishing realistic and reproducible evaluation protocols has become increasingly important.

\begin{wrapfigure}{r}{0.65\textwidth}
    \centering
    \includegraphics[width=0.68\textwidth]{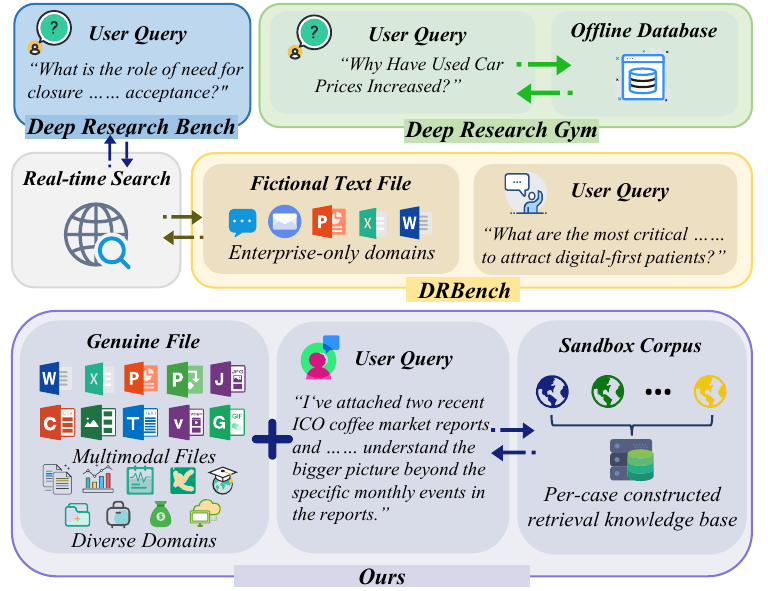}
    \caption{Comparison of deep research benchmarks. Given raw text queries, Deep Research Bench executes text queries via real-time search, and DeepResearchGym retrieves from a global offline database. \textbf{DRBench} incorporates user files (text modality) as input but relies on real-time search and focuses on the enterprise domain. In contrast, our \textbf{DR$^{3}$-Eval} processes both queries and files within a controlled sandbox corpus on diverse domains.}
    \label{fig_relationship}
\end{wrapfigure}

Evaluating deep research poses challenges that go beyond short-form reasoning or single-answer tasks. In realistic research settings, agents must infer implicit user intent from incomplete context, formulate effective search strategies, filter relevant evidence from large volumes of noise, and avoid hallucination while generating long-form reports~\citep{drbenchrealisticbenchmarkenterprise, liveresearchbenchlivebenchmarkusercentric, deerbenchmarkevaluatingdeep, sharma2025researchrubricsbenchmarkpromptsrubrics, deepresearchbenchcomprehensivebenchmark, mmdeepresearchbenchbenchmark, deepresearcheval, deepscholarbenchlivebenchmarkautomated, coelho2025deepresearchgym, li2026deepbench2, ruan2025expertlongbench}. As a result, evaluation outcomes are highly sensitive to the construction of tasks and information environments, revealing a fundamental tension between realism, controllability, and evaluability. Recently, as shown in Figure~\ref{fig_relationship}, several benchmarks have been proposed to evaluate deep research agents. For example, DeepResearch Bench ~\citep{deepresearchbenchcomprehensivebenchmark}~emphasizes open-ended report generation and long-horizon reasoning, but typically relies on live web access, making results difficult to reproduce and prone to evaluation ambiguity. DRBench ~\citep{drbenchrealisticbenchmarkenterprise}~improves structure by focusing on enterprise-style report generation from curated documents, yet largely omits explicit modeling of noisy or misleading information that is common in real-world research. More recently, DeepResearchGym ~\citep{coelho2025deepresearchgym}~introduces a sandbox-based framework that replaces live web access with fixed local corpora, improving reproducibility, but its tasks are purely text queries and lack grounding in authentic user research workflows.
%Ultimately, we present a set of practical guidelines for selecting RL tricks and composing effective combinations, tailored to various experimental settings. Empirical results show that applying our guidelines consistently outperforms existing strategies such as GRPO and DAPO, underscoring the practical utility and superior effectiveness of our approach.
%Ultimately, we hope that through these systematic empirical analyses, we can extract practical guidance principles and valuable takeaways that offer clear technical recommendations for researchers and engineers.    
As a result, \textit{there remains a gap between real-world research complexity—which involves multimodal user-provided materials, noisy and misleading information, and implicit research intent—and the environments used to evaluate DRA performance.}

% \begin{figure}[t]
% \centering
% \includegraphics[width=\textwidth]{figures/intro.pdf}
% \caption{Comparison of deep research benchmarks. 
% Given raw text queries,
% {Deep Research Bench} executes text queries via real-time search, and {DeepResearchGym} retrieves from a global offline database.
% \textbf{DRBench} incorporates user files (text modality) as input but relies on real-time search and focuses on the enterprise domain. In contrast, our \textbf{DR$^{3}$-Eval} processes both queries and files within a controlled sandbox corpus on diverse domains.}
% \vspace{-4mm}
% \label{fig_relationship}
% \end{figure}
To address these limitations, we introduce \textbf{DR$^{3}$-Eval}, a benchmark designed to reconcile realism, controllability, and reproducibility for deep research evaluation. DR$^{3}$-Eval targets report-generation tasks grounded in real user needs, constructed from authentic multimodal files that users have encountered in practice. 
Following the evaluation paradigm of DeepResearchGym~\citep{coelho2025deepresearchgym} and BrowseComp-Plus~\citep{browsecompplusfairtransparentevaluation}, we localize the retrieval corpus into a static, controlled sandbox rather than directly evaluating agents on the live web. Each task is paired with a per-case research sandbox corpus that simulates the open web while remaining fully static and verifiable. Within this sandbox, documents are carefully curated to include evidential sources, confounding documents, and ambient noise, enabling systematic analysis of an agent’s retrieval strategy, critical judgment, and robustness to distraction. Besides, a key feature of DR$^{3}$-Eval is its reverse-construction methodology: instead of posing open-ended questions with uncertain answerability, we derive each query from verified evidential documents, ensuring that every task admits a single, well-defined solution path. This design eliminates evaluation ambiguity while preserving the complexity of real research workflows. To support fine-grained assessment, we propose a multi-dimensional evaluation framework that measures Information Recall, Factual Accuracy, Citation Coverage, Instruction Following, and Depth Quality.
% combining deterministic checks with LLM-based expert judgment. 
% We further validate that our automated metrics strongly correlate with human expert evaluations.

Moreover, to demonstrate the utility of DR$^{3}$-Eval, we have developed \textbf{DR$^{3}$-Agent}, a multi-agent research system adapted to the benchmark’s closed-world setting, which can also take text user queries and corresponding user files (including text, image, video, audio, etc.) as input. Extensive experiments across state-of-the-art language models reveal that DR$^{3}$-Eval is highly challenging and exposes failure modes that are obscured by existing benchmarks.
% —most notably, the prevalence of hallucination under retrieval uncertainty and the non-trivial trade-offs between instruction adherence and factual correctness.

Overall, the contributions of our work are as follows:
\begin{itemize}
\item We propose DR$^{3}$-Eval, a realistic, reproducible, and multimodal benchmark for evaluating deep research agents for the report generation setting. 
\item We introduce a controlled sandbox-based task construction pipeline that balances real-world complexity with verifiable evaluation.

\item We provide comprehensive experimental analysis and diagnostics that offer new insights into the strengths and limitations of current LLMs as DRAs.
% .....multimodal understanding. encompassing not only traditional text documents such as PDFs but also a wide range of files, including structured tables, images, audio, video, and so on. 
\end{itemize}

\begin{figure*}[htbp]
\centering
\includegraphics[width=\textwidth]{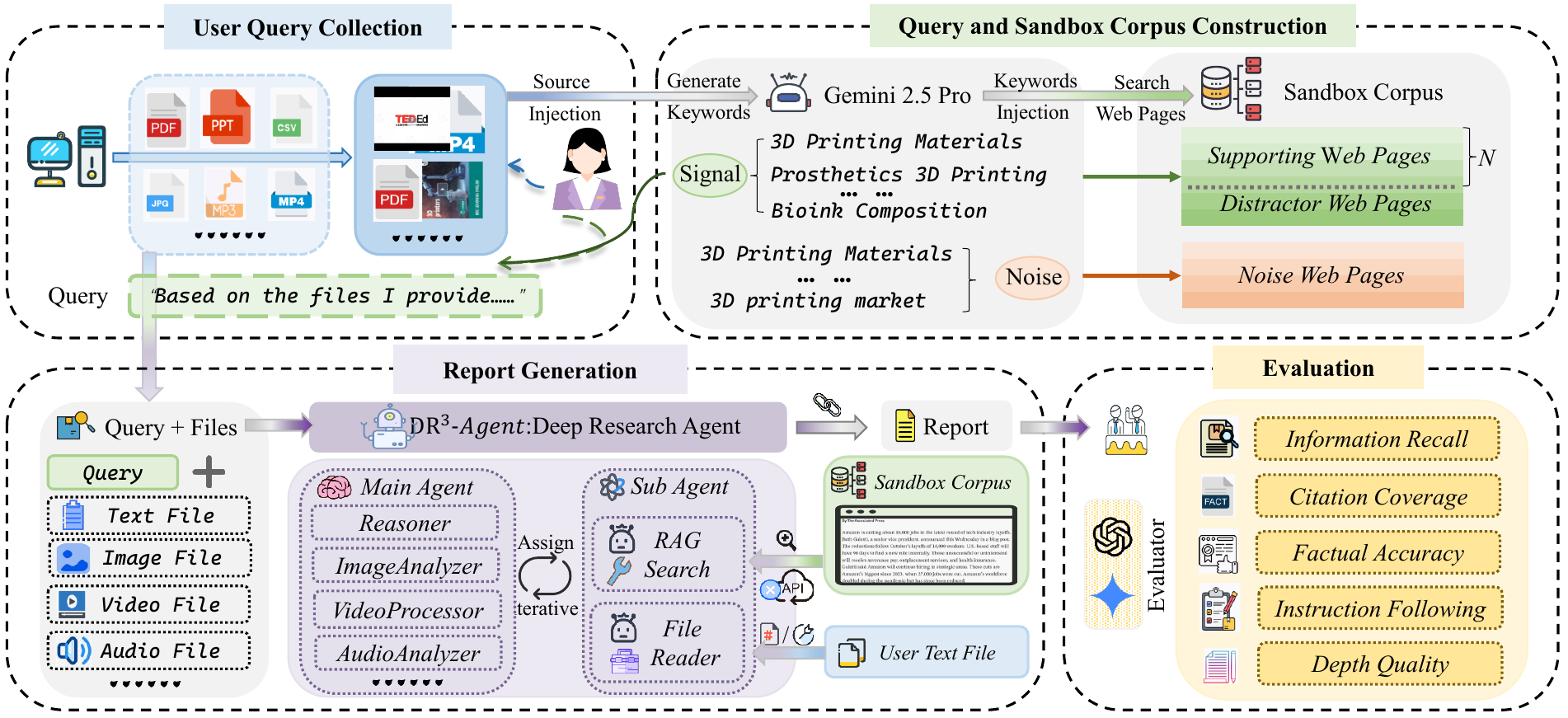}
\caption{Overview of the DR$^3$-Eval framework. (1) \textbf{Data construction} synthesizes search paths from real-world multimodal files via a divergent-convergent mechanism, establishing a static sandbox with controlled signal-to-noise ratios and backward-derived queries. (2) Our \textbf{DR$^3$-Agent} adopts a hierarchical multi-agent architecture where a perception-enhanced Main Agent coordinates global reasoning while specialized sub-agents execute iterative sandbox retrieval and file parsing. (3) \textbf{Evaluation protocol} utilizes a multidimensional metric suite to comprehensively assess performance in both evidence acquisition and analytical report generation.}
\label{fig_pipeline}
\end{figure*}

\section{Related Work}
\subsection{Deep Research Agent}
Deep Research Agents (DRAs) are specialized systems designed for complex, multi-stage research tasks. Their capabilities have evolved beyond the traditional question-answering paradigm~\citep{chen2026efficientmultimodalplanningagent}~ to autonomously planning long-horizon workflows, navigating heterogeneous web sources, and ultimately synthesizing information into structured, citation-grounded, expert-level reports~\citep{openai2025o3, google2025gemini, qwen2025deepresearch, perplexity2024research, grok1_2024,doubao2026}.
Currently, mainstream deep research systems have demonstrated powerful capabilities in handling complex research tasks, but they are mostly available as closed-source commercial products; open-source efforts, in contrast, emphasize modularity and reproducibility~ \citep{webthinkerempoweringlargereasoning,multimodaldeepresearchergeneratingtextchart,webresearcherunleashingunboundedreasoning, miroflow2025}. 
Nonetheless, a fundamental challenge remains: the reliance on live web environments for evaluation introduces uncontrollable temporal volatility.

\begin{table*}[t]
\caption{Comparison of our benchmarks with representative benchmarks. Columns report task type, covered domains, whether user files and sandbox corpus are supported, whether files are multimodal and in real-world scenarios, whether multiple files can be uploaded, and whether reverse construction is supported. Unlike prior work, DR$^3$-Eval combines user files and sandbox corpus, proposes a realistic, reproducible and multimodal benchmark for evaluating deep research agents in report-generation settings.}
\label{benchmark_comparison}
\centering
\resizebox{\textwidth}{!}{
\begin{tabular}{lcccccccc}
\toprule
Benchmark & Task type & Domain & User Files & Sandbox Corpus & Multi-modal & Real Scenario & Multi-File & Reverse Construction\\
\midrule
GAIA~\citep{gaiabenchmarkgeneralai}~ & QA & General & \cmark & \xmark & \cmark & \cmark & \xmark & \xmark\\
HLE ~\citep{phan2025humanitysexam}~ & QA & General & \cmark & \xmark & \cmark & \xmark & \cmark & \xmark\\
BrowseComp-Plus ~\citep{browsecompplusfairtransparentevaluation}~ & QA & General & \xmark & \cmark & \xmark & \cmark & \xmark & \cmark\\
DocBench ~\citep{docbenchbenchmarkevaluatingllmbased}~ & QA & General & \cmark & \xmark & \cmark & \cmark & \xmark & \xmark\\
MMLongBench-Doc \citep{mmlongbenchdocbenchmarkinglongcontextdocument}~ & QA & General & \cmark & \xmark & \cmark & \xmark & \xmark & \xmark\\
DRBench ~\citep{drbenchrealisticbenchmarkenterprise}~ & Report & Enterprise& \cmark & \cmark & \xmark & \cmark & \cmark & \cmark\\
Deep Research Bench ~\citep{deepresearchbenchcomprehensivebenchmark}~ & Report & General & \xmark & \xmark & \xmark & \cmark & \xmark & \xmark\\
Deep Scholar Bench ~\citep{deepscholarbenchlivebenchmarkautomated}~ & Report & Academic & \xmark & \xmark & \xmark & \cmark & \xmark  & \xmark\\
DEER ~\citep{deerbenchmarkevaluatingdeep}~ & Report & General & \xmark & \xmark & \xmark & \xmark & \xmark & \xmark\\
LiveResearch Bench ~\citep{liveresearchbenchlivebenchmarkusercentric}~ & Report & General & \xmark & \xmark & \xmark & \cmark & \xmark & \xmark\\
DeepResearchGym ~\citep{coelho2025deepresearchgym}~ & Report & General & \xmark & \cmark & \xmark & \cmark & \xmark  & \xmark\\
\midrule
\addlinespace
DR$^{3}$-Eval (Ours) & Report & General & \cmark & \cmark & \cmark & \cmark &\cmark & \cmark\\
\bottomrule
\end{tabular}
}
\end{table*}

\subsection{Deep Research Benchmark}
% TODO:多引用一些文章
The rapid development of Deep Research Systems has spurred the creation of numerous benchmarks designed to evaluate their diverse capabilities~\citep{xu2025comprehensivesurveydeepresearch}. Early efforts primarily addressed general reasoning and tool-use in QA scenarios~\citep{gaiabenchmarkgeneralai, phan2025humanitysexam},
which have recently evolved into complex information-seeking tasks in open-web environments~\citep{wei2025browsecomp}.
% TODO：deepwidesearch/widesearch
%which has also been extended to multimodal scenarios \citep{mmlongbenchdocbenchmarkinglongcontextdocument, docbenchbenchmarkevaluatingllmbased}.
%
% Reflecting the field's evolution, the benchmark's focus has also shifted from single-answer correctness to the comprehensive quality of the final report~\cite{deepresearcheval, deepresearchbenchiidiagnosing, deerbenchmarkevaluatingdeep,mmdeepresearchbenchbenchmark}.The open-ended nature of reports makes traditional metrics inadequate, and a methodology leveraging LLMs as judges has gained prominence, where these models are employed to assess fine-grained metrics.
% TODO：找4-5篇llm-as-a-judge的文章（别的地方引用过的文章也可以重复引用）——qingheng
% This evolution in task complexity has prompted a corresponding shift in evaluation paradigms, moving from the correctness of single answers to a more holistic assessment of final reports~\citep{deepresearcheval, deepresearchbenchiidiagnosing, deerbenchmarkevaluatingdeep,mmdeepresearchbenchbenchmark}. The open-ended nature of such reports renders traditional metrics insufficient, necessitating the use of LLMs as judges to evaluate fine-grained metrics.
However, a fundamental tension exists between reproducibility and realism in current evaluation environments. Benchmarks relying on live web access~ \citep{deepresearcheval, han2025deer}~ provide high ecological validity but suffer from temporal volatility, where fluctuating search results make performance comparisons inconsistent over time. Conversely, existing sandbox-based or local-corpus benchmarks ~\citep{coelho2025deepresearchgym, browsecompplusfairtransparentevaluation}~ ensure stability but often simplify the research context to ``clean'' and text-only data. They largely omit the multimodal complexity \citep{mmlongbenchdocbenchmarkinglongcontextdocument, docbenchbenchmarkevaluatingllmbased} and the confounding noise (outdated or biased information) inherent in authentic research.
Furthermore, as tasks shift from single-answer to report generation ~\citep{li2026deepbench2, mmdeepresearchbenchbenchmark}, traditional metrics have proven inadequate. This has necessitated the adoption of LLM-as-a-judge frameworks~\citep{liu2023geval, zheng2023judging, kim2023prometheus, zhu2023judgelm, docbenchbenchmarkevaluatingllmbased}~ to provide fine-grained, human-aligned assessments. Despite these advancements, there remains a gap for a benchmark that provides multimodal grounding, a noise-intensive yet static sandbox, and a verifiable solution path. 
In Table \ref{benchmark_comparison}, DR$^3$-Eval aims to address the above limitations by reconciling real-world research complexity with a rigorous, reproducible evaluation protocol.

\subsection{Data Construction}
Our dataset construction process involves five stages, designed to systematically create a benchmark for deep research that is grounded in real-world needs, features a controllable process, and enables precise evaluation.
Detailed prompts are provided in Appendix~\ref{app:data_prompts}.

\textbf{Stage 1: Grounding in Real-World Needs.}
Real-world research often requires synthesizing information across diverse data formats. To emulate this process, we recruited a group of paid volunteers, primarily comprising undergraduate and graduate students from various academic disciplines to ensure the breadth of our collected materials. 
They were tasked with providing intrinsically relevant material sets, whose compositions are inherently multimodal, encompassing text, structured data, static visuals, and dynamic media.
This process yielded 100 such document sets, evenly divided into 50 English and 50 Chinese sets. The topics cover three major domains—Technology, Economy, and Humanities—further broken down into 13 representative sub-fields, such as Computer Science, Healthcare, Finance, and Education. All collected materials then underwent a rigorous two-stage sanitization protocol: an automated script first identified and redacted personally identifiable information (PII), followed by a manual cross-validation by a separate group of annotators to ensure the complete anonymization of all personal, commercial, or proprietary data.

\textbf{Stage 2: Distilling Search Paths.}
Inspired by the strategy of generating distracting information via query expansion in BrowseComp-Plus~\citep{browsecompplusfairtransparentevaluation}~, we designed a two-stage ``divergent-convergent'' process~\citep{designcouncil2005,yao2023tree} 
to generate search keywords. The core of this design is to first broadly explore various aspects of a topic, and then precisely construct a solution path and its accompanying distractors from that exploration.
First, in the divergent stage, we leverage Gemini-2.5-Pro to perform an open-ended analysis of the source file, generating an initial set of 10 candidate keywords. The goal of this stage is to produce keywords that are conceptually diverse to cover different facets of the topic, thereby simulating a wide-ranging "brainstorming" session. Subsequently, in the convergent stage, the model evaluates this initial set and divides it into two categories: (1) Signal Keywords: which collectively point toward the core solution path; and (2) Noise Keywords: which are thematically related but designed to lead to irrelevant or misleading information.
Through this ``divergent-convergent'' process, we expand the evaluation challenge from simple information retrieval to the earlier and more advanced cognitive skills of query strategy formulation and path planning.

\textbf{Stage 3: Building Research Sandbox.}
To ensure the reproducibility of our evaluation and avoid potential cross-task interference, we construct a fully independent, static sandbox corpus for each task. The process begins by using the keywords from the previous stage to retrieve up to 100 web results for each keyword. After deduplicating all returned URLs, we employ a unified crawling and cleaning pipeline that filters out failed or erroneous pages and removes template elements such as navigation bars and ads.
Previously, benchmarks for web pages only distinguished between ``relevant'' and ``irrelevant''.
%
% However, a real-world deep research challenge comes from discerning seemingly relevant but low-quality information.
However, the challenge of deep research does not come from random noise alone, but from distinguishing genuinely useful evidence from seemingly relevant yet misleading information.
Therefore, we categorize all processed documents into three types, expanding upon the classification of retrieval quality proposed in CRAG~\citep{yang2024crag, yoran2023making}:
(1) Supportive Web Pages: high-relevance results from signal keywords, whose content is manually verified to provide necessary and sufficient evidence to answer the query; (2) Distractor Web Pages: also from signal keywords, but their content is confirmed to be outdated, one-sided, or inaccurate; (3) Noise Web Pages: results from noise keywords, used to systematically create evaluation environments with varying signal-to-noise ratios. The distribution is detailed in Appendix\ref{sec:webpage_distribution}.
To simulate the ``long-tail effect'' of information quality in real-world deep research, we design a fine-grained difficulty scaling strategy to construct evaluation settings with five different context lengths: 32k, 64k, 128k, 256k, and 512k tokens. To ensure the completeness and accessibility of the core solution path, all settings include the complete set of supportive web pages. Additionally, the number of distractor web pages increases proportionally with the total context length, and the remaining token quota is filled with noise web pages to reach the target length. During construction, these three categories of web pages are shuffled and randomly mixed.

\textbf{Stage 4: Constructing Query.} 
In report generation tasks, the open-ended nature of queries poses a significant challenge to objective, automated evaluation. Therefore, inspired by the backward design approach used in works like BrowseComp~\citep{wei2025browsecomp} for building QA benchmarks, we adopt an evidence-based, reverse construction method: we synthesize the final query based on the pre-determined evidential documents, integrated with the signal keywords. 
% This approach ensures that each query not only has a definitive, verifiable answer that is fully grounded within the research sandbox, but also that its solution path is unique and aligns with our predefined one.
This approach ensures that each query not only has a definitive, verifiable answer fully grounded in the sandbox, but also requires joint reasoning over the user files and specific web evidence, rather than being solvable through a single-step public search.

\textbf{Stage 5: Quality Control.}
We enforce a four-dimensional validation protocol to guarantee the rigorousness of candidate queries: (1) Implicit Guidance: queries must guide agents toward signal keywords without verbatim disclosure, preventing direct information leakage. (2) Synthesis Necessity: a ``leave-one-out'' verification is employed to ensure that the final answer is strictly contingent on combining the initial user files with specific web evidence; candidate tasks are discarded if their core conclusion can be directly obtained through single-step public search. (3) Insight Novelty: queries are disqualified if the core factual claims of the golden insight are directly retrievable from public search engines, thereby blocking shortcut solutions and preserving the need for cross-source reasoning. (4) Interpretative Unambiguity: each query undergoes manual inspection to eliminate ambiguity and guarantee a singular, precise interpretation.

We further summarize this filtering process as a QC funnel. Starting from 280 candidate tasks collected from volunteers, 105 were discarded during the leave-one-out validation stage due to multiple plausible interpretations or the inability to derive a unique solution path within the sandbox, and another 75 were filtered out because their factual difficulty was insufficient. The final benchmark therefore contains 100 tasks, corresponding to a pass rate of 35.7\%, yielding a high-purity and low-ambiguity task set.

% \begin{itemize}
%     \item \textbf{Implicit Guidance:} Queries must guide agents toward Signal Keywords without verbatim disclosure, preventing direct information leakage.
%     \item \textbf{Synthesis Necessity:} We employ a ``leave-one-out'' verification to ensure the final answer is strictly contingent on combining the initial user file with specific web evidence.
%     \item \textbf{Insight Novelty:} Queries are disqualified if the core factual claims of the golden insight are directly retrievable from public search engines, ensuring the necessity of reasoning.
%     \item \textbf{Interpretative Unambiguity:} Each query undergoes manual inspection to eliminate ambiguity and guarantee a singular, precise interpretation.
% \end{itemize}
\begin{figure*}[t]
\centering
\includegraphics[width=\textwidth]{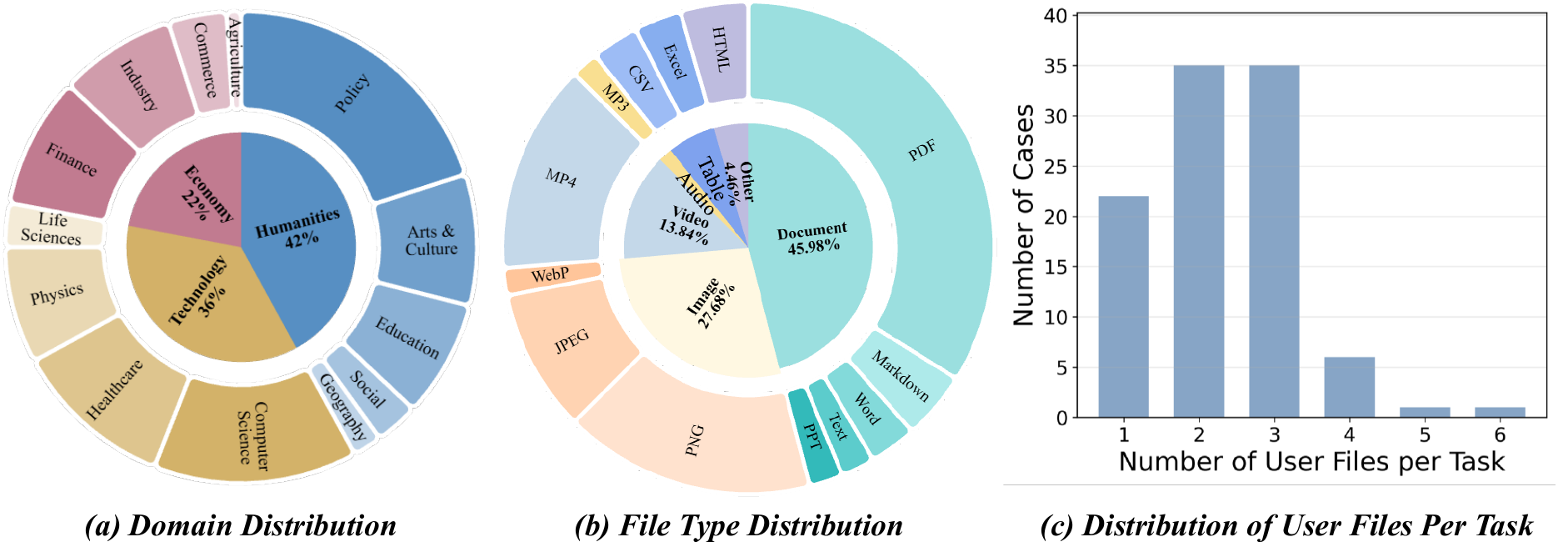}
\caption{Dataset statistics. 
(a) Domain coverage spanning Technology, Economy, and Humanities, comprising 13 atomic sub-domains. 
(b) Distribution of file types. 
(c) Distribution of user files per task.}
\label{datasets_stat}
\end{figure*}

\subsection{Dataset Statistics}
Through rigorous manual curation, DR$^3$-Eval comprises 100 independent tasks with an even split between English and Chinese samples. As shown in Figure \ref{datasets_stat}(a), the tasks cover technology, economy, and humanities, subdivided into 13 atomic domains such as computer science, healthcare, and policy. Regarding input modalities, Figure \ref{datasets_stat}(b) presents a distribution of 45.98\% documents, 27.68\% images, and 13.84\% videos, alongside data sheets, audio, and HTML files, with specific format breakdowns detailed in Appendix \ref{Detailed breakdown of file types}. These inputs involve significant data scale, with 68\% of tasks being multi-modal, where PDFs average 11.21 pages, Excel spreadsheets contain 215.14 rows, and videos last 3 minutes and 27 seconds. Figure \ref{datasets_stat}(c) further plots the file density per task, which averages 2.24 user files and reaches a maximum of 6. Complementing these internal user resources to establish a realistic evaluation environment, the sandbox corpus introduces massive external noise, containing an average of 465.5 web pages per task under the 512k token configuration.

\section{DR$^{3}$-Agent}
\label{DR-Agent}

\subsection{Framework Construction}
% TODO:tongyi_deepresearch_2025, deerflow_2025, camel_workforce_2026
To address the deep research tasks of DR$^3$-Eval involving User Files and a Sandbox Corpus, we develop DR$^3$-Agent, an LLM-driven system based on the MiroFlow framework~\citep{miromindteam2025mirothinkerpushingperformanceboundaries}. It is worth emphasizing that current open-source deep research frameworks (e.g., DeerFlow~\citep{deerflow_2025}, Qwen-DeepResearch~\citep{tongyi_deepresearch_2025}, and Camel-workforce~\citep{camel_workforce_2026}) typically cannot directly handle the offline closed sandbox environment and the cross-reading of multimodal files tasks proposed in DR$^3$-Eval. Therefore, as illustrated in Figure~\ref{fig_pipeline}, we integrate perception tools directly into the main agent to effectively handle multimodal user files such as audio and video. This design enables it to synthesize video and audio content within the global context, rather than treating them as isolated extraction tasks. Supported by these perception capabilities and a built-in Python execution environment, the main agent serves as the system's reasoning hub. It maintains the global task context and runs a dynamic ``Plan-Act-Observe'' loop to formulate action plans and coordinate sub-agents for specific information acquisition tasks.

At the information acquisition level, to mitigate the main agent's context burden, the system employs two dedicated sub-agents powered by the same underlying LLM. While sharing the model backbone, these sub-agents do not share the global state and return only highly condensed summaries to the main agent. Specifically, the RAG search sub-agent interacts with the static sandbox corpus. 
We replace the original open web search with an iterative dense retrieval mechanism based on text-embedding-3-small, employing the ReAct~\citep{yao2022react} paradigm within a controlled environment to refine queries and perform multiple, continuous iterative retrievals within the sandbox corpus. Unlike conventional RAG systems, which typically rely on a separate retriever to fetch top-$k$ chunks from a static knowledge base, our RAG sub-agent performs autonomous multi-step retrieval with iterative query refinement. This requires the agent to evaluate incomplete or conflicting evidence and revise its search direction across iterations, making the search process functionally analogous to heuristic exploration over hyperlink graphs. 
Meanwhile, the file reader sub-agent specializes in parsing long-text user files, utilizing tools to execute fine-grained keyword queries and retrieve content by page numbers.

\subsection{Evaluation Metrics}

DR$^3$-Eval comprises five complementary metrics, categorized into two dimensions: \textit{Information Seeking}, which assesses the quality of gathered evidence, and \textit{Report Generation}, which evaluates the final output quality. Among these, for the four metrics requiring semantic assessment, we utilize $\Phi$ (GPT-5.1) as the evaluator, with specific prompts detailed in Appendix~\ref{evaluation prompt}.

\subsubsection{Information Seeking}

\textbf{Information Recall (IR)}
% 引用ok 
% We employ Gemini-2.5-Flash to extract insight sets $\mathcal{I}_\text{UF}$ and $\mathcal{I}_\text{SC}$ \citep{nuggetrecall, gingernugget}, from user files and the sandbox corpus, respectively. Subsequently, we use the evaluator model $\Phi$ to assess the report $R$'s coverage of each insight $i$, assigning a score $\text{cov}(i, R) \in \{1, 0.5, 0\}$. IR calculates the ratio of strictly fully covered insights\citep{deepscholarbenchlivebenchmarkautomated}.
We employ Gemini-2.5-Flash to extract insight sets $\mathcal{I}_\text{UF}$ and $\mathcal{I}_\text{SC}$ ~\citep{nuggetrecall, gingernugget}~ from user files and the sandbox corpus, respectively, using prompts detailed in Appendix \ref{Insights Extraction from User Files} and \ref{Insights Extraction from Sandbox Corpus}. All extracted insights are manually verified to ensure accuracy. Subsequently, we use the evaluator model $\Phi$ to assess the report $R$'s coverage of each insight $i$, assigning a score $\text{cov}(i, R) \in \{1, 0.5, 0\}$. IR calculates the ratio of strictly fully covered insights~\citep{deepscholarbenchlivebenchmarkautomated}.

\begin{align}
\label{eq:inforecall_uf}
\text{IR}_\text{UF}(R, \mathcal{I}_\text{UF}) &= \frac{1}{|\mathcal{I}_\text{UF}|} \sum_{i \in \mathcal{I}_\text{UF}} \mathbb{1}[\text{cov}(i, R) = 1] \\
\label{eq:inforecall_sc}
\text{IR}_\text{SC}(R, \mathcal{I}_\text{SC}) &= \frac{1}{|\mathcal{I}_\text{SC}|} \sum_{i \in \mathcal{I}_\text{SC}} \mathbb{1}[\text{cov}(i, R) = 1]
\end{align}

% \textbf{Citation Coverage (CC)}
% Inspired by the irreplaceable literature metric from Deep Scholar Bench~\citep{deepscholarbenchlivebenchmarkautomated}, we establish the ground truth set $\mathcal{D}_{\text{req}}$, comprising user files and supportive web pages strictly necessary for the query. Let $\mathcal{D}_{\text{cited}}$ denote the documents explicitly cited in $R$. The coverage is defined as:
\textbf{Citation Coverage (CC)}
Inspired by the irreplaceable literature metric from DeepScholar-Bench~\citep{deepscholarbenchlivebenchmarkautomated}, we establish the ground truth set $\mathcal{D}_{\text{req}}$, comprising user files and supportive web pages strictly necessary for the query. Strongly tied to our overall reverse construction process, this metric evaluates the macroscopic ``information gathering recall,'' thereby reflecting the model's research-oriented retrieval ability. Let $\mathcal{D}_{\text{cited}}$ denote the documents explicitly cited in $R$. The coverage is defined as:

\begin{equation}
\label{eq:citcov}
\text{CC}(R, \mathcal{D}_{\text{req}}) = \frac{|\mathcal{D}_{\text{req}} \cap \mathcal{D}_{\text{cited}}|}{|\mathcal{D}_{\text{req}}|}
\end{equation}

\subsubsection{Report Generation}

% 这个不引用了
% TODO：这里还要改成来自于userfile 和 sandbox corpus两方面源头的引用
% We extract the set of all claim-source pairs $\mathcal{C}$ from $R$. For each pair $(c, s) \in \mathcal{C}$, we define an indicator function $\mathbb{I}(c, s)$ that equals 1 if the source $s$ textually entails the claim $c$, and 0 otherwise:
% \begin{equation}
% \label{eq:factacc}
% \text{FA}(R) = \frac{1}{|\mathcal{C}|} \sum_{(c, s) \in \mathcal{C}} \mathbb{I}(c, s)
% \end{equation}
\textbf{Factual Accuracy (FA)}
We extract the set of all claim-source pairs $\mathcal{C}$ from the generated report $R$. To ensure robust verification across modalities, we employ $\Phi$ to evaluate textual claims, while utilizing Gemini-2.5-Pro to verify claims grounded in video or audio content. The source $s$ originates from either the user files or the sandbox corpus. For each pair $(c, s) \in \mathcal{C}$, we define an entailment function $\mathbb{V}(c, s)$ that equals 1 if $s$ supports $c$, and 0 otherwise:
\begin{equation}
\label{eq:factacc}
\text{FA}(R) = \frac{1}{|\mathcal{C}|} \sum_{(c, s) \in \mathcal{C}} \mathbb{V}(c, s)
\end{equation}

\textbf{Instruction Following (IF)}
% 引用ok
% We utilize $\Phi$ to generate a checklist $\mathcal{L}$ based on the query, covering aspects such as content, evidence, and analysis\citep{sharma2025researchrubricsbenchmarkpromptsrubrics, liveresearchbenchlivebenchmarkusercentric}. We then evaluate whether $R$ satisfies each requirement $l \in \mathcal{L}$, indicated by a binary satisfaction score $\mathbb{S}(l, R) \in \{1, 0\}$:
We utilize $\Phi$ to generate a checklist $\mathcal{L}$ based on the task query (prompt in Appendix \ref{Checklist Generation}), covering aspects such as content, evidence, and analysis~\citep{sharma2025researchrubricsbenchmarkpromptsrubrics, liveresearchbenchlivebenchmarkusercentric}. All checklists are manually verified. We then evaluate whether $R$ satisfies each requirement $l \in \mathcal{L}$, indicated by a binary satisfaction score $\mathbb{S}(l, R) \in \{1, 0\}$:
\begin{equation}
\label{eq:if}
\text{IF}(R, \mathcal{L}) = \frac{1}{|\mathcal{L}|} \sum_{l \in \mathcal{L}} \mathbb{S}(l, R)
\end{equation}

\textbf{Depth Quality (DQ)} 
% 不引用了
We employ the model $\Phi$ as an expert judge to evaluate the analytical substance and logical rigor of $R$. The quality score is assigned conditioned on the query $Q$ and a predefined rubric $\mathcal{P}$:
\begin{equation}
\label{eq:depthqual}
\text{DQ}(R, Q) = \Phi(R, Q \mid \mathcal{P})
\end{equation}

A sample report generated by DR$^3$-Agent is provided in Appendix \ref{sec:report_example}, with its evaluation details in Appendix \ref{sec:evaluation_example}.

\section{Experiments}

\subsection{Experimental Settings}
% 
% To systematically evaluate the overall capabilities of large language models under
For DR$^{3}$-Agent, the maximum interaction turns for the main agent is set to 10, while the sub-agents for RAG and file reading are limited to 5 and 3 turns, respectively.
% In the ablation study, the number of turns for the sub-agent browsing is 10. 
We utilize OpenAI's text-embedding-3-small for vectorization.
For baselines, 
we evaluate GPT-4.1~\citep{GPT-4.1modelcard}, Claude Sonnet 4~\citep{ClaudeSonnet4modelcard}, Gemini 2.5 Pro~\citep{Gemini2.5Promodelcard}, Qwen3-235B-A22B~\citep{Qwen3-235B-A22Bmodelcard}, Qwen3-30B-A3B~\citep{Qwen3-30B-A3Bmodelcard}, Qwen3-32B~\citep{Qwen3-32Bmodelcard}, GLM-4.6~\citep{GLM-4.6modelcard}~ and GLM-4.7~\citep{GLM-4.7.modelcard}.
% Qwen3-8B,
% Besides, we set the temperature of all models to 0.3. 
In the evaluation phase, 
for text modality,
we introduce GPT-5.1~\citep{GPT-5.1modelcard}~ as the judge model, and for multimodal contents (e.g., audio and video), we use Gemini-2.5-Pro as an assistant judge.
To ensure the evaluation is deterministic, the temperature for all judge models is set to 0.
Additional runtime and API cost statistics are provided in Appendix \ref{cost}.

\subsection{Main Results}

\begin{table*}[th]
\centering
\small % 缩小字体以适应页面宽度
\caption{Evaluation results on DR$^{3}$-Agent. The best and second-best performances are highlighted in \textbf{bold} and \underline{underlined}, respectively. \textbf{Key}: $\text{IR}_{UF}$/$\text{IR}_{SC}$ = Information Recall from user files/sandbox corpus; \textbf{CC} = Citation Coverage; \textbf{FA} = Factual Accuracy; \textbf{IF} = Instruction Following; \textbf{DQ} = Depth Quality; \textbf{Avg.} = Average.}
\resizebox{\textwidth}{!}{%
\begin{tabular}{l 
    *{21}{>{\centering\arraybackslash}p{0.48cm}}}
\toprule
\multirow{3}{*}{\textbf{Models}} 
& \multicolumn{9}{c}{\textbf{Information Seeking}} 
& \multicolumn{9}{c}{\textbf{Report Generation}} 
& \multicolumn{3}{c}{\textbf{Total Score}} \\
\cmidrule(lr){2-10} \cmidrule(lr){11-19} \cmidrule(lr){20-22}
& \multicolumn{3}{c}{$\text{IR}_{UF}$} 
& \multicolumn{3}{c}{$\text{IR}_{SC}$} 
& \multicolumn{3}{c}{\textbf{CC}} 
& \multicolumn{3}{c}{\textbf{FA}} 
& \multicolumn{3}{c}{\textbf{IF}} 
& \multicolumn{3}{c}{\textbf{DQ}} 
& \multicolumn{3}{c}{\textbf{Avg. }} \\
\cmidrule(lr){2-4} \cmidrule(lr){5-7} \cmidrule(lr){8-10} \cmidrule(lr){11-13} \cmidrule(lr){14-16} \cmidrule(lr){17-19} \cmidrule(lr){20-22}
& 64k & 128k & 512k & 64k & 128k & 512k & 64k & 128k & 512k & 64k & 128k & 512k & 64k & 128k & 512k & 64k & 128k & 512k & 64k & 128k & 512k \\
\midrule
Claude Sonnet 4 & \textbf{58.8} & \textbf{60.4} & \textbf{60.8} & \textbf{55.3} & \underline{46.6} & \textbf{41.8} & \underline{64.7} & \underline{54.8} & \textbf{48.5} & \textbf{87.0} & \textbf{82.7} & 82.1 & \underline{87.4} & \underline{89.2} & \textbf{88.5} & \underline{70.7} & \underline{71.5} & \underline{72.0} & \textbf{70.7} & \textbf{67.5} & \textbf{65.6} \\
GLM-4.7 & \underline{55.7} & \underline{55.0} & \underline{57.1} & \underline{53.1} & \textbf{47.6} & \underline{42.1} & \textbf{65.4} & \textbf{55.9} & 45.3 & 84.5 & 82.1 & 80.3 & \textbf{88.8} & \textbf{89.3} & 88.1 & \textbf{71.1} & \textbf{71.8} & \textbf{72.1} & \underline{69.8} & \underline{66.9} & \underline{64.1} \\
GLM-4.6  & 53.4 & 52.6 & 50.3 & \underline{49.5} & 43.9 & 39.8 & 58.2 & 52.0 & 44.0 & 84.0 & \underline{82.3} & \underline{82.9} & 85.6 & 87.2 & 86.4 & 70.1 & 69.3 & 70.6 & 66.8 & 64.5 & 62.3 \\
Gemini-2.5-Pro  & 43.9 & 45.7 & 42.9 & 37.7 & 35.1 & 30.8 & 54.3 & 49.5 & 36.6 & 81.3 & 80.7 & 80.0 & 84.9 & 84.5 & 84.5 & 67.1 & 68.3 & 67.4 & 61.5 & 60.6 & 57.0 \\
GPT-4.1 & 40.7 & 42.5 & 41.3 & 30.9 & 29.4 & 29.2 & 37.2 & 35.6 & 30.0 & 56.4 & 54.2 & 58.8 & 83.0 & 83.3 & 82.7 & 63.1 & 63.2 & 63.4 & 51.9 & 51.3 & 50.9 \\
Qwen3-235B-A22B  & 37.4 & 36.0 & 39.7 & 35.7 & 29.8 & 28.8 & 40.6 & 36.6 & 31.8 & 52.5 & 53.6 & 49.8 & 78.0 & 78.6 & 80.2 & 62.1 & 62.8 & 61.9 & 51.1 & 49.6 & 48.7 \\
Qwen3-32B & 33.2 & 36.6 & 35.4 & 26.5 & 25.3 & 24.7 & 34.2 & 32.3 & 26.1 & 49.4 & 52.2 & 51.5 & 73.5 & 74.2 & 74.3 & 58.8 & 59.9 & 59.3 & 45.9 & 46.7 & 45.2 \\
% TODO 512k 64k事实准确性太高
% Qwen3-8B & 27.7 & 32.9 & 29.9 & 27.8 & 25.1 & 22.7 & 36.2 & 38.2 & 29.7 & 57.9 & 44.3 & 60.0 & 69.0 & 73.6 & 67.6 & 56.3 & 58.1 & 57.4 & 45.8 & 45.4 & 44.6 \\
Qwen3-30B-A3B & 30.9 & 38.2 & 34.1 & 23.2 & 25.7 & 23.5 & 26.6 & 25.3 & 21.5 & 41.9 & 46.8 & 45.2 & 73.2 & 71.8 & 74.7 & 57.6 & 58.2 & 58.0 & 42.2 & 44.3 & 42.8 \\
% Gemini-2.5-Flash & 50.7 & 52.7 & 51.8 & 42.9 & 37.0 & 34.8 & 48.8 & 44.8 & 38.6 & \underline{86.1} & \underline{82.3} & 81.2 & 78.8 & 81.0 & 82.2 & 66.1 & 67.7 & 67.8 & 62.2 & 60.9 & 59.4 \\

\bottomrule
\end{tabular}}
% \caption{Evaluation results on DR$^{3}$-Agent. The best and second-best performances are highlighted in \textbf{bold} and \underline{underlined}, respectively.}

\label{tab:mainresults}
\end{table*}

% \begin{figure}[t]
% \centering
% \includegraphics[width=\textwidth]{figures/heatmap_no8b.pdf}
% % \caption{Performance of cross-disciplinary for different models.}
% \caption{Performance of different LLMs across different domains. }
% \label{fig_heatmap}
% \vspace{-5mm}
% \end{figure}

\begin{wrapfigure}{r}{0.55\textwidth}
    \centering
    % \vspace{-2mm}
    \includegraphics[width=\linewidth]{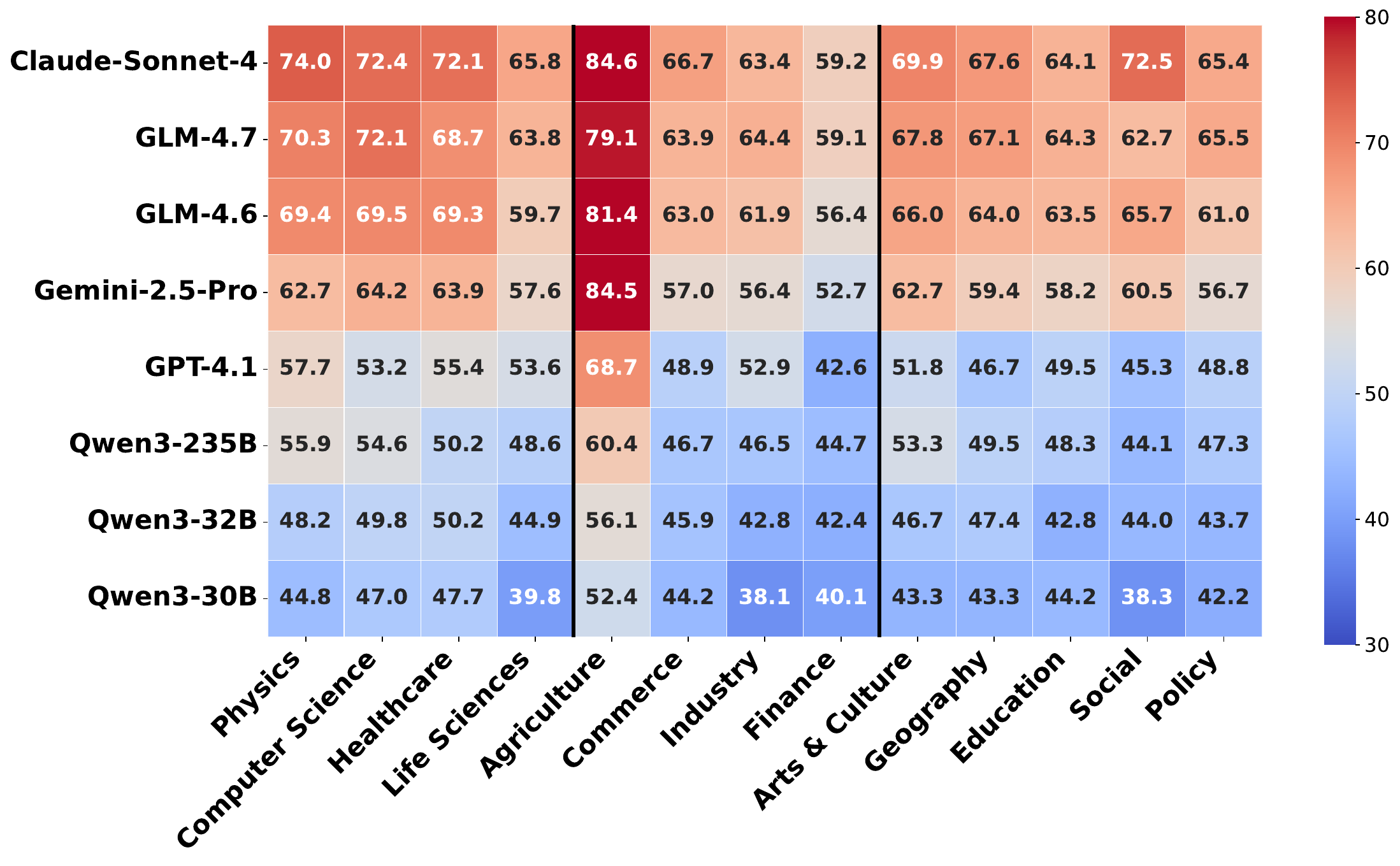}
    \caption{Performance of different LLMs across different domains.}
    \label{fig_heatmap}
    \vspace{-10mm}
\end{wrapfigure}

In Table \ref{tab:mainresults} and Figure~\ref{fig_heatmap}, we provide the results of different models, and have the following observations:
% our leads to the following core conclusions:
(1) \textbf{DR$^{3}$-Eval is very challenging. }
% Different performance levels among models in our DR$^{3}$-Eval. 
Claude Sonnet 4  achieves the best results. 
Besides, within the same model family (i.e., Qwen), the scaling law is still a key factor in complex tasks.
(2) \textbf{Longer contexts lead to lower performance.}
As the size of the sandbox corpus grows from 64k to 512k, a general drop in performance is seen across all models. 
We suppose that the longer contexts result in noisier and more irrelevant contexts, which make it more difficult to obtain valuable insights.
(3) \textbf{Better instruction following does not indicate higher factual accuracy.}
For example, some models (e.g., Qwen3-235B-A22B and GPT-4.1) achieve relatively good results in Instruction Following(IF), but obtain very low factual accuracies.
% even when their performance on both Information Recall (IR) is average.
This suggests these models cannot accurately obtain sufficient information from the given materials: they tend to create a report that ``looks'' complete and satisfies the query, at the high cost of Factual Accuracy (FA). 
(4) \textbf{Performance varies a lot across different domains and different models}. In Figure~\ref{fig_heatmap}, we observe that results from several domains (e.g., GLM 4.7 achieves the best on ``Industry'' domains while Claude Sonnet 4 achieves the best on ``Physics'').

% Therefore, the large gap in FA has become a key factor for telling models' abilities apart.
% \textbf{Models have a hard time using external knowledge bases, showing a clear source preference.}

% Models generally achieve low scores on the Information Recall (IR) metrics, which highlights the challenging nature of our benchmark.
% %
% Information Recall from User File($\text{IR}_{UF}$) is consistently better than Information from Sandbox Corpus ($\text{IR}_{SC}$). This ``source preference'' shown when processing different information sources reveals that models are not robust or efficient when using external knowledge bases that are less structured, more complex, and contain noise.
% Figure\ref{fig_heatmap} shows the domain-level performance of the DR$3$-Agent. For each domain, we randomly select 3 reports from that domain for scoring visualization. It can be seen that in relatively difficult tasks, the performance of agents generally declines, such as Public Policy \& Law.
 
% Even the best-performing model, Claude Sonnet 4, the average score(Avg) drops from 70.7 to 65.6. 
%
% Keeping up high performance in complex and noisy real-world research settings is still a common problem.

\subsection{Further Analysis}

\begin{wrapfigure}{r}{0.40\textwidth}
    \centering
    \includegraphics[width=\linewidth]{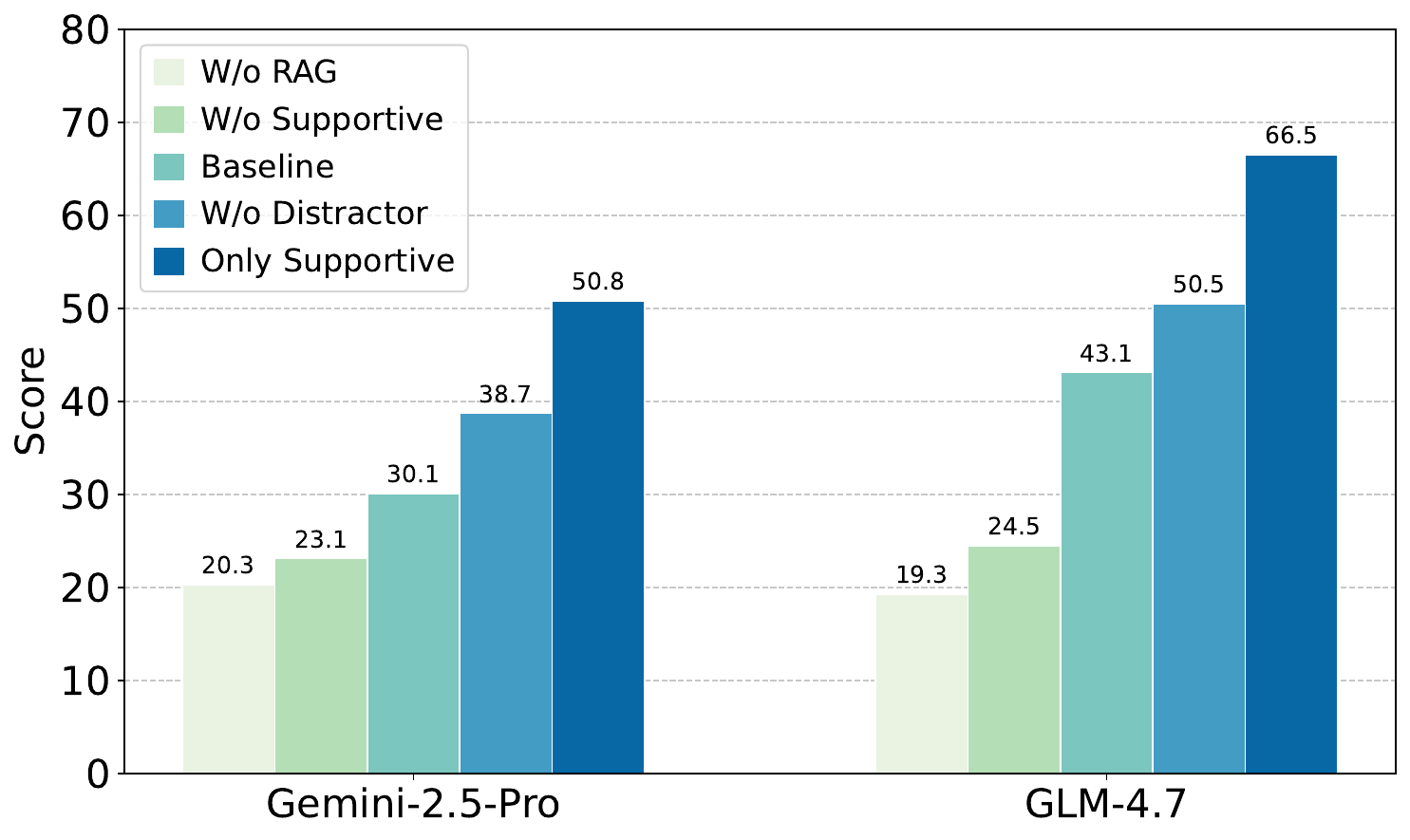}
    \caption{Analysis on the effectiveness of sandbox corpus.}
    \label{ablation_longcontext}
\end{wrapfigure}
\paragraph{Evaluation stability and significance.} A 10,000-iteration bootstrap analysis shows no overlap in the 95\% confidence intervals between the top two models, with a Wilcoxon test ($p=0.0046$) confirming a significant difference in their scores. Furthermore, the total score variance across repeated evaluations is only 0.874, and the Kendall's $\tau$ and Spearman's $\rho$ for model rankings under resampling reach 0.969 and 0.991, respectively.
Taking Claude Sonnet 4, GLM-4.7, and GLM-4.6 as examples, the standard deviations across three repeated runs for each model remain exceptionally low at 0.83, 0.85, and 1.33, respectively.

\paragraph{Analysis on the correlation between sandbox corpus and real-world web corpus. }

To further verify whether the sandbox corpus can approximate information acquisition in real-world web environments, we conduct experiments with real-time web search on an English subset using Qwen3-235B and Gemini-2.5-Pro. 
As shown in Table \ref{tab:ablation_compact}, the overall performance remains close across the two settings, with particularly high consistency in Citation Coverage. This indicates that the core evidence chains ultimately relied upon by the models in the live web setting highly overlap with the supportive documents predefined in the sandbox. Overall, no obvious systematic bias is observed between the local sandbox and real-time web search, suggesting that the sandbox preserves the main information difficulty that determines task performance and can serve as a reliable substitute for web retrieval.

\paragraph{Analysis on the performance of different sizes of sandbox corpus. }
\begin{wraptable}{r}{0.52\textwidth}
    \centering
    \caption{Analysis on the correlation between sandbox corpus and real-world web corpus.}
    \label{tab:ablation_compact}
    \small
    \setlength{\tabcolsep}{3pt}
    \resizebox{\linewidth}{!}{
    \begin{tabular}{@{}lcccccc@{}}
    \toprule
    & \multicolumn{3}{c}{\textbf{Qwen3-235B-A22B}} & \multicolumn{3}{c}{\textbf{Gemini-2.5-Pro}} \\
    \cmidrule(lr){2-4} \cmidrule(lr){5-7}
    \textbf{Metric} & Baseline & w/ Web & Change ($\Delta$) & Baseline & w/ Web & Change ($\Delta$) \\
    \midrule
    $\text{IR}_{SC}$ & 33.2 & 38.5 & (+5.3) & 40.4 & 41.9 & (+1.5) \\
    $\text{IR}_{UF}$ & 23.9 & 20.2 & (-3.7) & 27.4 & 25.4 & (-2.0) \\
    CC & 36.3 & 28.0 & (-8.3) & 50.4 & 49.0 & (-1.4) \\
    FA & 59.0 & 60.3 & (+1.3) & 76.3 & 75.9 & (-0.4) \\
    IF & 73.6 & 79.2 & (+5.6) & 80.1 & 84.1 & (+4.0) \\
    DQ & 63.8 & 62.0 & (-1.8) & 67.8 & 70.4 & (+2.6) \\
    \textbf{Avg.} & 48.3 & 48.0 & (-0.3) & 57.1 & 57.8 & (+0.7) \\
    \bottomrule
    \vspace{-5pt}
    \end{tabular}
    }
\end{wraptable}
% \begin{wraptable}{r}{0.52\textwidth}
%     \centering
%     \caption{Analysis on the correlation between sandbox corpus and real-world web corpus.}
%     \label{tab:ablation_compact}
%     \small
%     \setlength{\tabcolsep}{3pt}
%     \resizebox{\linewidth}{!}{
%     \begin{tabular}{@{}lcccccc@{}}
%     \toprule
%     & \multicolumn{3}{c}{\textbf{Qwen3-235B-A22B}} & \multicolumn{3}{c}{\textbf{Gemini-2.5-Pro}} \\
%     \cmidrule(lr){2-4} \cmidrule(lr){5-7}
%     \textbf{Metric} & Baseline & w/ Web & Change ($\Delta$) & Baseline & w/ Web & Change ($\Delta$) \\
%     \midrule
%     $\text{IR}_{SC}$ & 31.0 & 28.1 & (-2.8) & 32.8 & 34.9 & (+2.1) \\
%     $\text{IR}_{UF}$ & 43.6 & 47.9 & (+4.3) & 36.9 & 42.4 & (+5.4) \\
%     FA & 63.3 & 65.9 & (+2.6) & 75.6 & 70.2 & (-5.4) \\
%     DQ & 67.0 & 65.5 & (-1.5) & 69.5 & 72.0 & (+2.5) \\
%     IF & 73.4 & 79.1 & (+5.7) & 81.4 & 80.4 & (-1.0) \\
%     \textbf{Avg.} & 54.9 & 56.6 & (+1.6) & 63.1 & 62.1 & (-1.0) \\
%     \bottomrule
%     \end{tabular}
%     }
% \end{wraptable}

\begin{figure*}[!t]
\centering
\includegraphics[width=\textwidth]{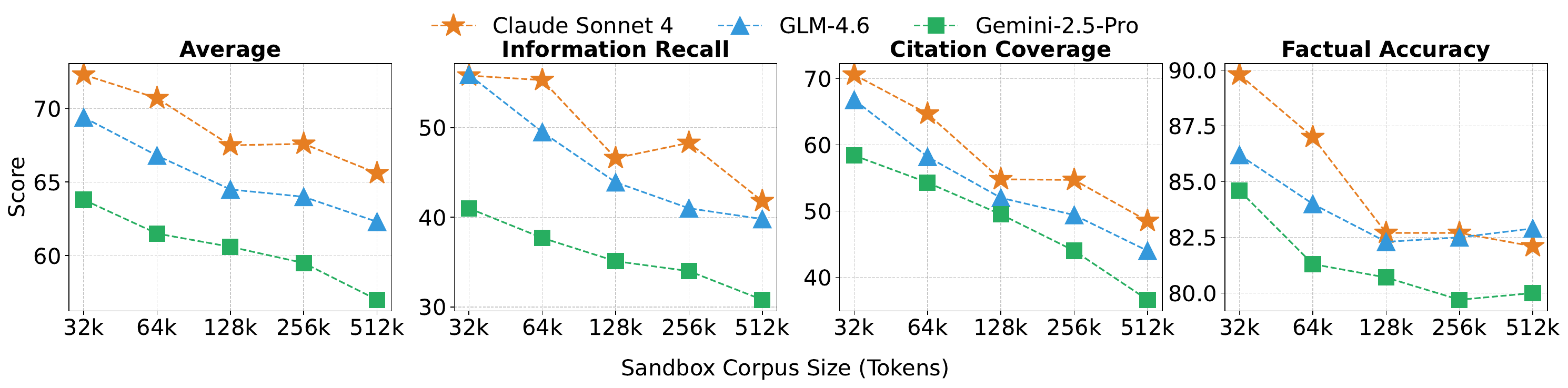}
% \caption{Performance of four models is evaluated on the average score (AVG) and three sub-metrics ($\text{IR}_{SC}$, CC, FA) as context length of sandbox corpus increases from 32k to 512k tokens.}
\caption{Analysis on the performance of different sizes of sandbox corpus. }
\label{fig_ablation_scale}
\end{figure*}

We conduct evaluations on five sandbox corpora  (i.e., 32k, 64k, 128k, 256k, and 512k tokens). As shown in Figure \ref{fig_ablation_scale}, as the size of the sandbox corpus increases, the overall performance (Avg.), $\text{IR}_{SC}$ and CC of all models shows a clear downward trend. This shows that the models not only find it hard to locate relevant evidence chunks, but also find it harder to identify effective answers among the increasing noise and distracting information.
However, FA metric shows relative stability. We believe this mainly measures the model's ability to reason correctly after obtaining the relevant information, which is closer to its inherent performance. But as the size of sandbox corpora grows, the retrieved text snippets may contain more noise or fail to contain the correct answer due to retrieval failure (i.e., a low $\text{IR}_{SC}$ score), which result in lower scores.
% increasing the possibility of producing hallucinations.
% As shown in Figure\ref{fig_ablation_scale}, 

% \begin{wrapfigure}{r}{0.30\columnwidth}
%     \centering
%     \vspace{-3mm}
%     \includegraphics[width=\linewidth]{figures/dr3_vs_deerflow.pdf}
%     \caption{Comparison of framework architectures.}
%     \label{deerflow}
% \end{wrapfigure}

\paragraph{Comparison of framework architectures.}
To further validate the effectiveness of our framework, we conduct a comparative experiment with DeerFlow on a subset. Considering that DeerFlow's native retrieval mechanism cannot directly process DR$^3$-Eval, we transplant the Agentic RAG component from DR$^3$-Agent to ensure a fair evaluation. 

\begin{wrapfigure}{r}{0.40\columnwidth}
    \centering
    % \vspace{-3mm}
    \includegraphics[width=\linewidth]{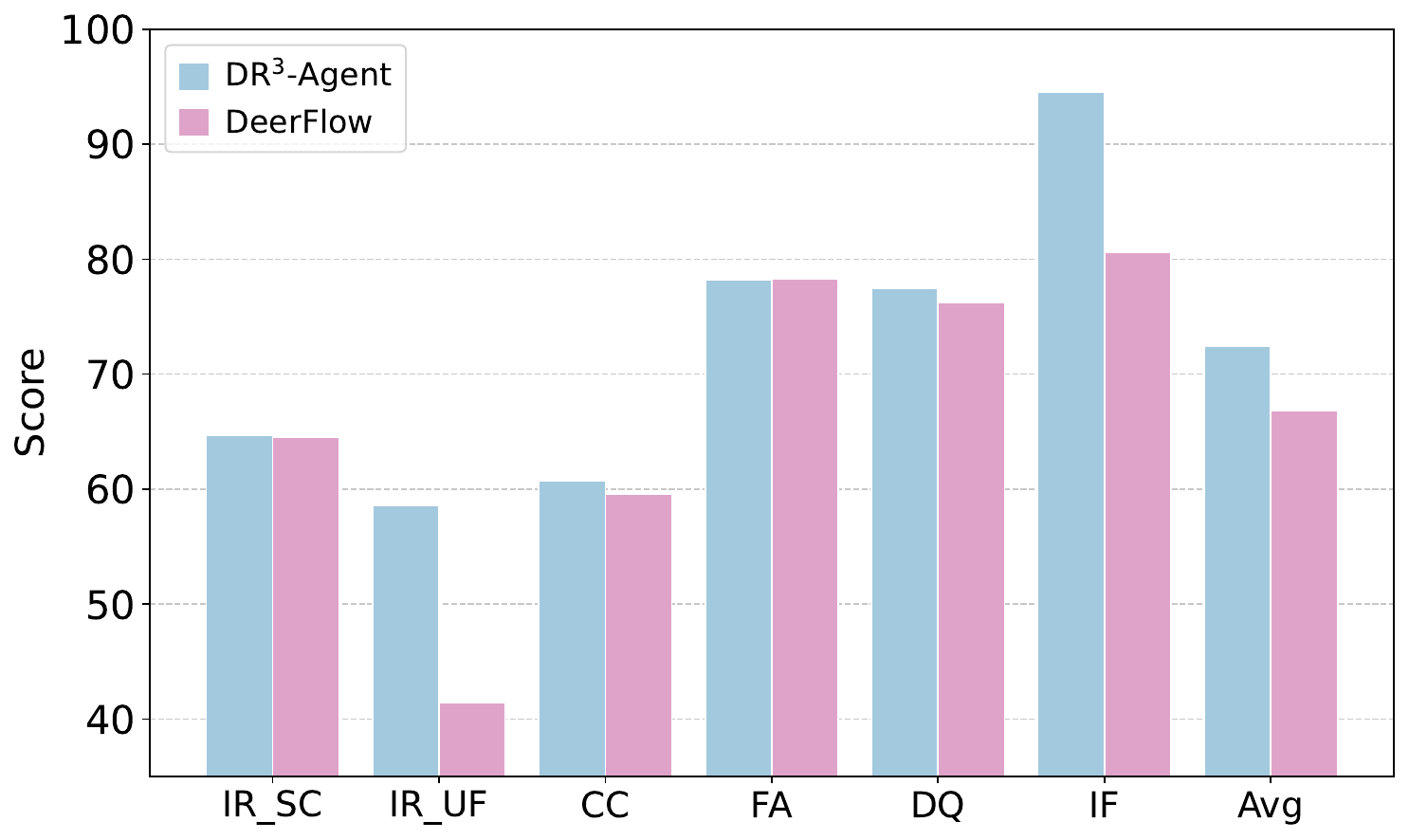}
    \caption{Comparison of framework architectures.}
    \label{deerflow}
\end{wrapfigure}

As shown in Fig \ref{deerflow}, equipped with the same component, both frameworks exhibit converging capabilities in basic information acquisition when processing standard continuous long texts. 
However, DR$^3$-Agent demonstrates distinct architectural advantages when tackling complex deep research tasks. 
Information within user files is typically more fragmented than in standard documents, while DR$^3$-Agent can more stably integrate these discrete pieces of evidence. Besides, DR$^3$-Agent consistently adheres to task instructions even under information overload.

\paragraph{Analysis on the effectiveness of sandbox corpus. }
% To verify the reasonableness of our sandbox corpus design, we systematically analyze the impact of different document components on model performance. 
To verify the reasonableness of our sandbox corpus design, we systematically analyze the impact of different document components on model performance using a sample of 20 tasks.
The experiments are mainly based on the 128k-sized corpus (except for the only supportive setting). In Figure \ref{ablation_longcontext}, after removing the distractor web pages, the performance of all models improved significantly. This shows that our distractor documents effectively increased the task difficulty. 
Furthermore, we observe that the agent's performance is nearly identical when the sandbox corpus is provided without supportive documents compared to when no corpus is provided at all. This demonstrates that, apart from the designated supportive documents, our sandbox corpus contains no other effective information that the agent can exploit to complete the task. When only supportive web pages are present in the sandbox corpus, it establishes the model's performance upper bound under perfect information retrieval.

\begin{wraptable}{r}{0.4\columnwidth}
    \centering
    \small
    \caption{LLM-as-judge vs. human evaluation.}
    \label{tab:comparison}
    \setlength{\tabcolsep}{4pt}
    \begin{tabular}{lccc}
    \toprule
    \textbf{Method} & \textbf{$r$} & \textbf{$\rho$} & \textbf{Agr.} \\
    \midrule
    DR$^{3}$-Eval (Ours) & 0.78 & 0.73 & 0.89 \\
    Inter-Human & 0.83 & 0.76 & 0.91 \\
    \bottomrule
    \end{tabular}
\end{wraptable}

\paragraph{Analysis on the correlation between LLM-as-judge and human evaluation. }
% To validate the alignment between DR$^{3}$-Eval and human judgment, we conducted a correlation study on 30 reports randomly sampled across all domains, independently reviewed by four experts to yield 60 evaluation samples. Consistency was measured using Pearson and Spearman correlation coefficients~ \citep{deerbenchmarkevaluatingdeep} ~alongside pairwise agreement ~\citep{deepresearchbenchcomprehensivebenchmark}~, with calculation details provided in Appendix \ref{sec:human_evaluation}. In Table \ref{tab:comparison}, our automated scoring exhibits strong concurrence with expert evaluations, notably achieving a pairwise agreement exceeding 90\%.
To validate the alignment between DR$^{3}$-Eval and human judgment, we conducted a correlation study on 50 reports randomly sampled across all domains, which were independently reviewed by four experts. Consistency was measured using Pearson and Spearman correlation coefficients~\citep{deerbenchmarkevaluatingdeep}~alongside pairwise agreement~\citep{deepresearchbenchcomprehensivebenchmark}~, with calculation details provided in Appendix \ref{sec:human_evaluation}. As shown in Table \ref{tab:comparison}, our automated scoring exhibits strong concurrence with expert evaluations. Furthermore, we evaluated the consistency between our automated claim extraction for factual accuracy and human annotations, achieving a Precision of 0.924 and a Recall of 0.960.

% \paragraph{Analysis on the effectiveness of different judge LLMs. }
% To verify the reliability of different judge models, we select GPT-5.1, Gemini-2.5-Pro and Qwen-Max as alternative judges to re-rank and evaluate six models using sampled 30 reports.
% % each on the same task.
% %
% % To verify the robustness of our evaluation framework and the reliability of different judge models, we select GPT-5.1, Gemini-2.5-Pro and Qwen-Max as alternative judges to re-rank and evaluate six models. 
% The rankings from Gemini-2.5-Pro and Qwen-Max yielded Spearman's rank correlations of \textbf{0.886} and \textbf{0.943}, respectively, when compared against the rankings from GPT-5.1. Details are shown in Appendix~\ref{app:ranking_analysis}. We believe this slight fluctuation in ranking is mainly due to the ``model bias'' phenomenon, where judge models tend to give higher scores to models from their own series. For example, Qwen-Max ranked its own series model, Qwen3-235B-A22B, ahead of GPT-4.1, while the other judges' rankings were the opposite. 

\begin{wraptable}{r}{0.5\columnwidth}
    \vspace{-11mm}
    \centering
    \small
\caption{Effectiveness of different retrievers.}
\label{tab:retrieval_comparison}
\setlength{\tabcolsep}{3.5pt}
    \begin{tabular}{lccc}
    \toprule
    \textbf{Model} & \textbf{OpenAI-Emb} & \textbf{Qwen-Emb} & \textbf{BM25} \\
    \midrule
    GLM-4.7        & \textbf{56.58} & \textbf{53.61} & \textbf{50.71} \\
    GPT-4.1        & 36.15          & 35.64          & 22.60          \\
    Gemini-2.5-Pro & 49.51          & 37.16          & 31.25          \\
    \bottomrule
    \end{tabular}
\end{wraptable}

\paragraph{Analysis on the effectiveness of different judge LLMs. }
To verify the reliability of different judge models, we select Claude Sonnet 4, Gemini-2.5-Pro, and Qwen-Max as alternatives to GPT-5.1 to re-rank and evaluate six specific models. Compared against the rankings from GPT-5.1, the rankings derived from their scores are almost identical, with a mean Spearman's $\rho$ of 0.924. Details are shown in Appendix~\ref{app:ranking_analysis}. We believe this slight fluctuation in ranking is mainly due to the ``model bias'' phenomenon, where judge models tend to give higher scores to models from their own series. Regarding the multimodal assistant judge LLM, Gemini-2.5-Pro, we replace it with Qwen3-VL-Plus and Kimi-k2, yielding a mean Spearman's $\rho$ of 0.864. The impact on the final scores is not significant, with an average difference of less than 2 points ($p > 0.05$), demonstrating that the scoring is highly robust.

% TODO：这里缺了表格（还得再整理一下数据）
\paragraph{Effect of maximum iteration turns in Agentic-RAG.}
We conduct an ablation study on the maximum iteration turns of RAG. Considering that our RAG sub-agent employs a ReAct-based Agentic-RAG rather than a traditional single-shot Top-$K$ retrieval, we set the maximum iteration turns to 1, 3, 5, and 7. Shown in Table \ref{tab:rag_max_turns}, as the allowed iteration turns increase, the overall performance of different models exhibits a clear upward trend, particularly in IR and CC. However, we observe that once the iteration turns increase to a certain extent, the model's performance not only peaks but even experiences a slight decline. 

%

%
% We infer that even when equipped with powerful retrieval capabilities, the agent inevitably falls into cognitive overload when continuously accumulating massive context history or coping with extremely dense adversarial noise (e.g., up to 512k tokens of distractor pages).

\begin{wraptable}{r}{0.5\columnwidth}
    \vspace{-4mm}
    \centering
    \small
    \caption{Effect of different maximum RAG turns on IR and CC.}
\label{tab:rag_max_turns}
    \setlength{\tabcolsep}{3.5pt}
    \begin{tabular}{lcccc}
    \toprule
    & \multicolumn{2}{c}{\textbf{Qwen3-235B-A22B}} & \multicolumn{2}{c}{\textbf{Gemini-2.5-Pro}} \\
    \cmidrule(lr){2-3} \cmidrule(lr){4-5}
    \textbf{Turns} & \textbf{IR} & \textbf{CC} & \textbf{IR} & \textbf{CC} \\
    \midrule
    1 & 27.2 & 14.8 & 32.4 & 21.0 \\
    3 & 34.7 & 27.1 & 39.6 & 47.6 \\
    5 & 33.9 & 27.1 & \textbf{44.6} & \textbf{51.0} \\
    7 & \textbf{44.0} & \textbf{32.9} & 38.1 & 48.1 \\
    \bottomrule
    \end{tabular}
\end{wraptable}

\paragraph{Analysis on the effectiveness of different retrievers. }
We select three representative models and compare three different retrieval strategies on Citation Coverage (CC) using a 128k-sized corpus. By default, we use OpenAI text-embedding-3-small to build the vector database and compare its results with Qwen-text-embedding-v2 and BM25.
In Table \ref{tab:retrieval_comparison}, the text-embedding-3-small achieves the best performance on all models, while Qwen-text-embedding-v2 performs slightly lower than it. 
% This may be because the embedding space of text-embedding-3-small is more compatible with the internal semantic representations of the models in this evaluation (e.g., GLM, GPT, and Gemini). 
In contrast, the traditional lexical-based method, BM25, performs significantly worse.
% because it cannot handle the problem of semantic equivalence between queries and documents.

% \begin{table}[t]
% \caption{Analysis on the effectiveness of different retrievers. } 
% \label{tab:retrieval_comparison}
% \centering
% \small
% % \resizebox{\columnwidth}{!}{
% \begin{tabular}{lccc}
% \toprule
% \textbf{Model} & \textbf{OpenAI-Emb} & \textbf{Qwen-Emb} & \textbf{BM25} \\
% \midrule
% GLM-4.7          & \textbf{56.58} & \textbf{53.61} & \textbf{50.71} \\
% GPT-4.1          & 36.15 & 35.64 & 22.60 \\
% Gemini-2.5-Pro   & 49.51 & 37.16 & 31.25 \\
% % Qwen3-30B-A3B         & 25.75 & 21.64 & 17.53 \\
% \bottomrule
% \end{tabular}
% % }
% \vspace{-5mm}
% \end{table}

\paragraph{Case Study. }
% We conduct an error attribution analysis on five key models, classifying the root causes of errors into three categories: 
% We conduct an error attribution analysis on five selected models, based on a random sample of 20 reports per model, and classify the root causes into three categories:
% (1) \textbf{Retrieval Error}, denoting where the agent fails to locate or omits key information required to answer the question during the retrieval stage; 
% (2) \textbf{Reasoning Error}, denoting where the agent, despite obtaining relevant information, makes mistakes in information integration, logical inference, or detail processing; and (3) \textbf{Hallucination}, denoting where the model's generated response is not based on the provided context but is instead fabricated from its parametric knowledge.
% In Figure~\ref{case_study}, hallucination is the primary cause of failure for most models. This indicates that when faced with retrieval difficulties or ambiguous information, models tend to abandon their reliance on external knowledge and instead resort to their internal knowledge to make inaccurate ``guesses''. Among them, Gemini-2.5-Pro exhibits almost no reasoning errors, whereas Qwen3-235B has a low retrieval failure rate, yet its hallucination rate is the highest among all models. This reflects 
% that even when the model successfully retrieves information, it fails to effectively utilize these sources to generate the answer.
We conduct an error attribution analysis on five selected models, based on 100 reports per model, and classify the root causes into three categories:

\begin{wrapfigure}{r}{0.50\columnwidth}
    \centering
    \includegraphics[width=\linewidth]{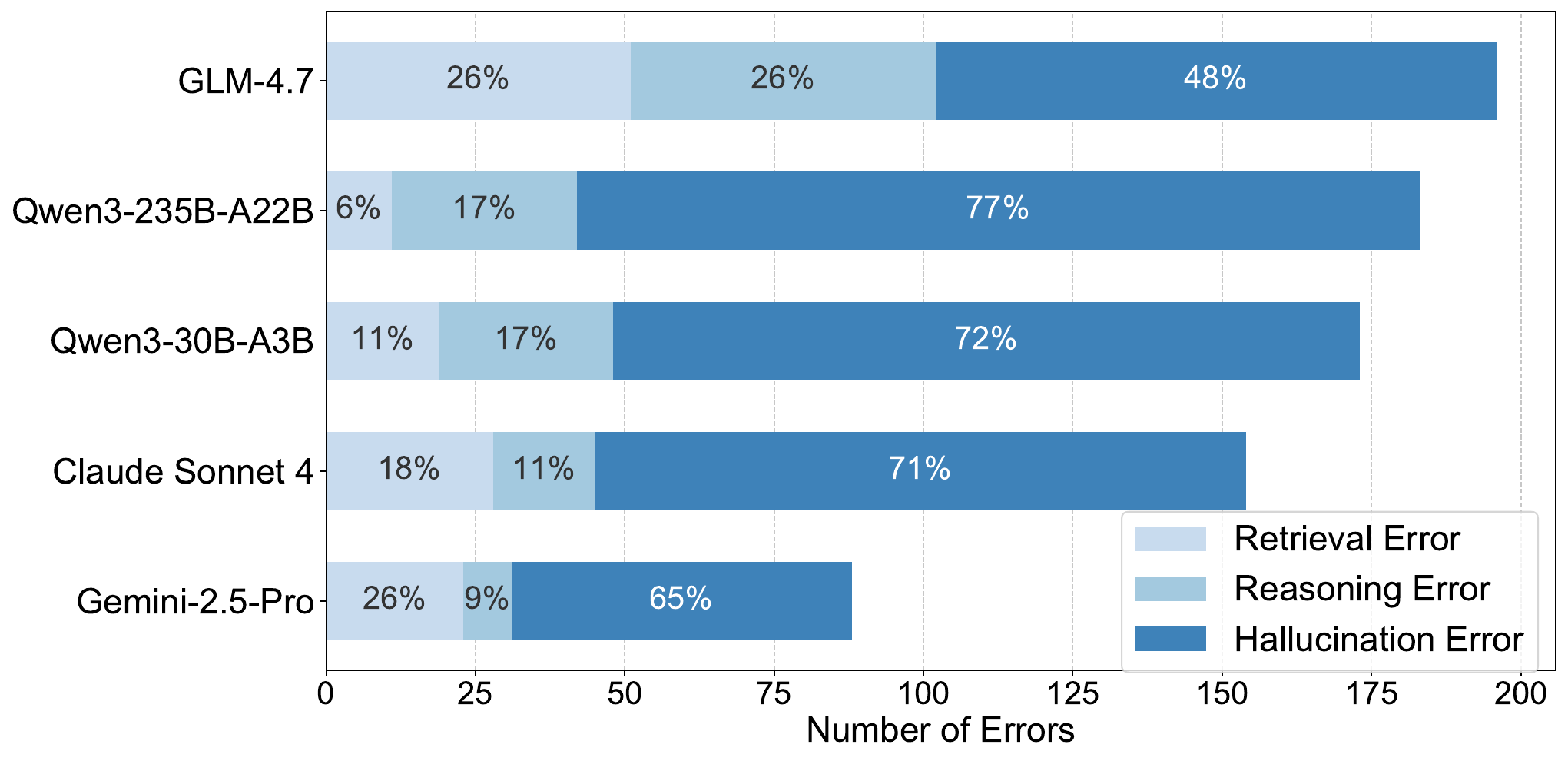}
    \caption{Error type analysis across LLMs.}
    \label{case_study}
\end{wrapfigure}

(1) \textbf{Retrieval Error}, denoting where the agent fails to locate or omits key information required to answer the question during the retrieval stage;
(2) \textbf{Reasoning Error}, denoting where the agent, despite obtaining relevant information, makes mistakes in information integration, logical inference, or detail processing; and
(3) \textbf{Hallucination}, denoting where the model's generated response is not based on the provided context but is instead fabricated from its parametric knowledge.
In Figure~\ref{case_study}, hallucination remains the primary cause of failure for most models. This indicates that in long-horizon research tasks, the key challenge for current models lies not only in whether they can retrieve relevant information, but also in whether they can remain grounded in external evidence when generating the final report. In contrast, the distributions of retrieval and reasoning errors vary across models: some tend to fail earlier at the evidence acquisition stage, while others still deviate from the evidence during subsequent integration and generation even after obtaining relevant information. Overall, these results suggest that the main bottleneck of current models lies in the stability of evidence utilization, rather than in evidence acquisition alone.

% \begin{figure}[!t]
% \vspace{5mm} 
% \centering
% \includegraphics[width=\textwidth]{figures/error_types_font.pdf}
% \caption{Analysis of different error types for different LLMs.}
% \vspace{-6mm}
% \label{case_study}
% \end{figure}

% \begin{wrapfigure}{r}{0.48\textwidth}
%     \centering
%     \includegraphics[width=\linewidth]{figures/error_types_font.pdf}
%     \caption{Analysis of different error types for different LLMs.}
%     \label{case_study}
% \end{wrapfigure}

% \paragraph{Visualization}

\section{Conclusion}
This work introduces DR$^{3}$-Eval, a benchmark designed to address key limitations in the evaluation of deep research agents. By grounding tasks in authentic user research scenarios, constructing controlled yet web-like sandbox environments, and eliminating evaluation ambiguity through reverse task construction, DR$^{3}$-Eval provides a principled testbed for assessing long-horizon research capabilities. Our experimental results show that DR$^{3}$-Eval poses substantial challenges for state-of-the-art LLMs and reveals systematic failure modes of these LLMs.
% in retrieval robustness and hallucination control. that are not well captured by existing benchmarks. We hope DR$^{3}$-Eval will facilitate more reliable evaluation and drive the development of more robust and trustworthy deep research agents.

% DR$^{3}$-Agent is a comprehensive benchmark for evaluating the report generation capabilities of Deep Research Agents in a multimodal and multifile scenario. It features a multidimensional classification and hybrid evaluation framework. Our experiments demonstrate that,

% \section*{Impact Statements}

% This paper presents work whose goal is to advance the field of machine learning. Specifically, we introduce DR$^3$-Eval to benchmark and improve Deep Research Agents.

\section*{Impact Statements}

This paper presents DR$^{3}$-Eval, a benchmark designed to advance the evaluation of Deep Research Agents. Our work has several broader impacts and ethical considerations:

\textbf{Data Privacy and Human Subjects:} The dataset construction involved collecting authentic files from human participants. To address ethical concerns regarding privacy, all participants were compensated for their contributions. We implemented a rigorous two-stage sanitization protocol, comprising both automated redaction and manual cross-validation, to ensure that all Personally Identifiable Information (PII) and sensitive proprietary data were completely removed before inclusion in the benchmark.

\textbf{Societal Implications of Deep Research Agents:} The advancement of autonomous research agents holds significant potential to increase productivity in knowledge-intensive fields. However, we acknowledge the risks associated with this technology, such as the potential for generating convincing hallucinations or the misuse of automated information gathering for malicious purposes. By introducing metrics specifically focused on \textit{Factual Accuracy} and \textit{Citation Coverage}, and by providing a static, reproducible sandbox environment, our work aims to steer the field towards developing safer, more reliable, and verifiable systems, rather than those that merely optimize for persuasiveness.

\textbf{Environmental Impact:} Furthermore, by utilizing a static sandbox corpus rather than relying on repetitive live-web crawling for every evaluation run, our framework promotes more computationally efficient and environmentally sustainable benchmarking practices.

\bibliographystyle{unsrtnat}
\bibliography{references} 

\newpage
\appendix
\onecolumn
% Due to file size limitations, we are unable to include the complete dataset in the Supplementary Material. Instead, we provide a subset in the attachments to demonstrate the data quality and format. 

\section{Detailed breakdown of file types}
\label{Detailed breakdown of file types}

\begin{figure}[h]
\centering
\begin{subfigure}[b]{0.48\textwidth}
\centering
\includegraphics[width=\linewidth]{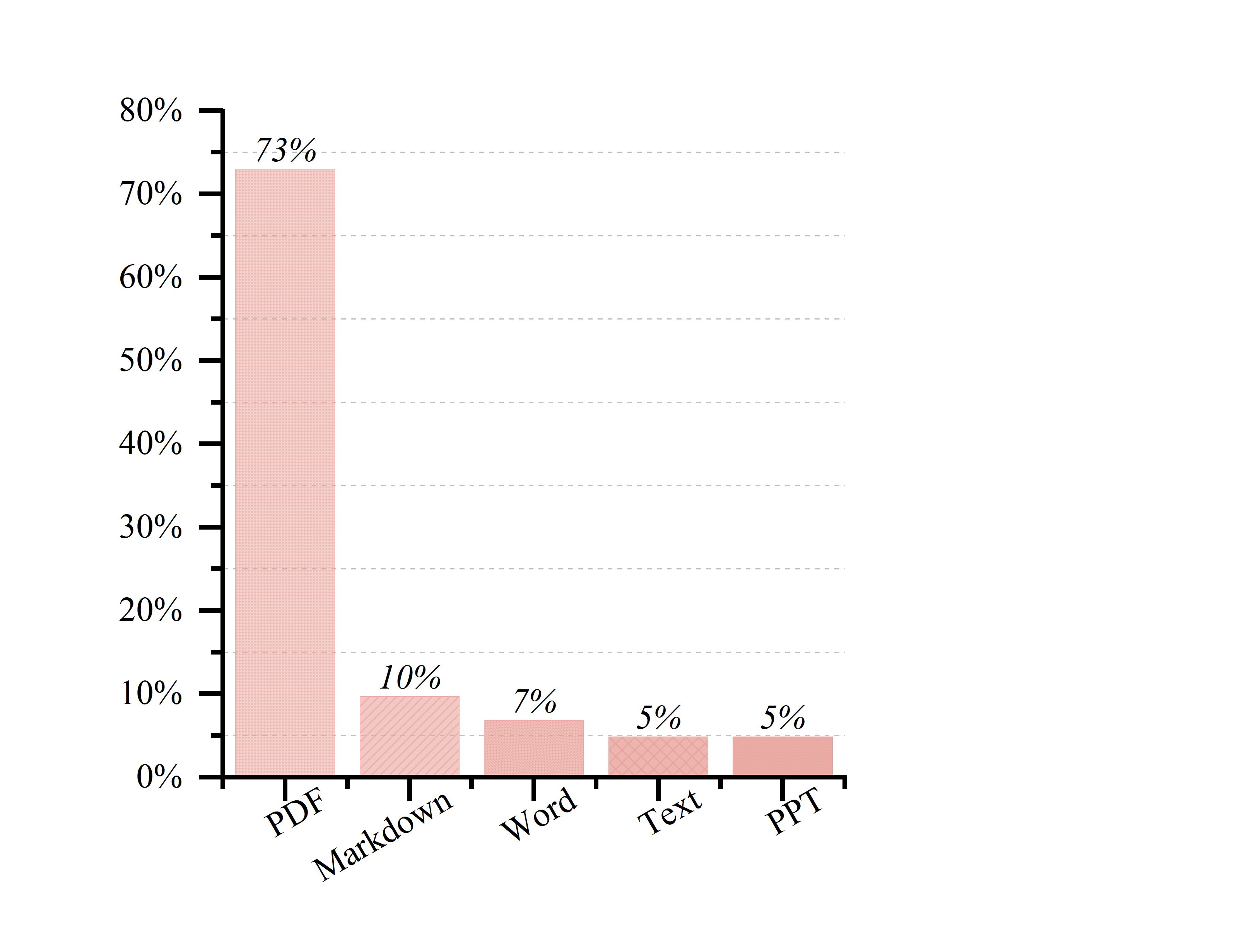} 
\caption{document formats}
\label{fig:doc_dist}
\end{subfigure}
\hfill
\begin{subfigure}[b]{0.48\textwidth}
\centering
\includegraphics[width=\linewidth]{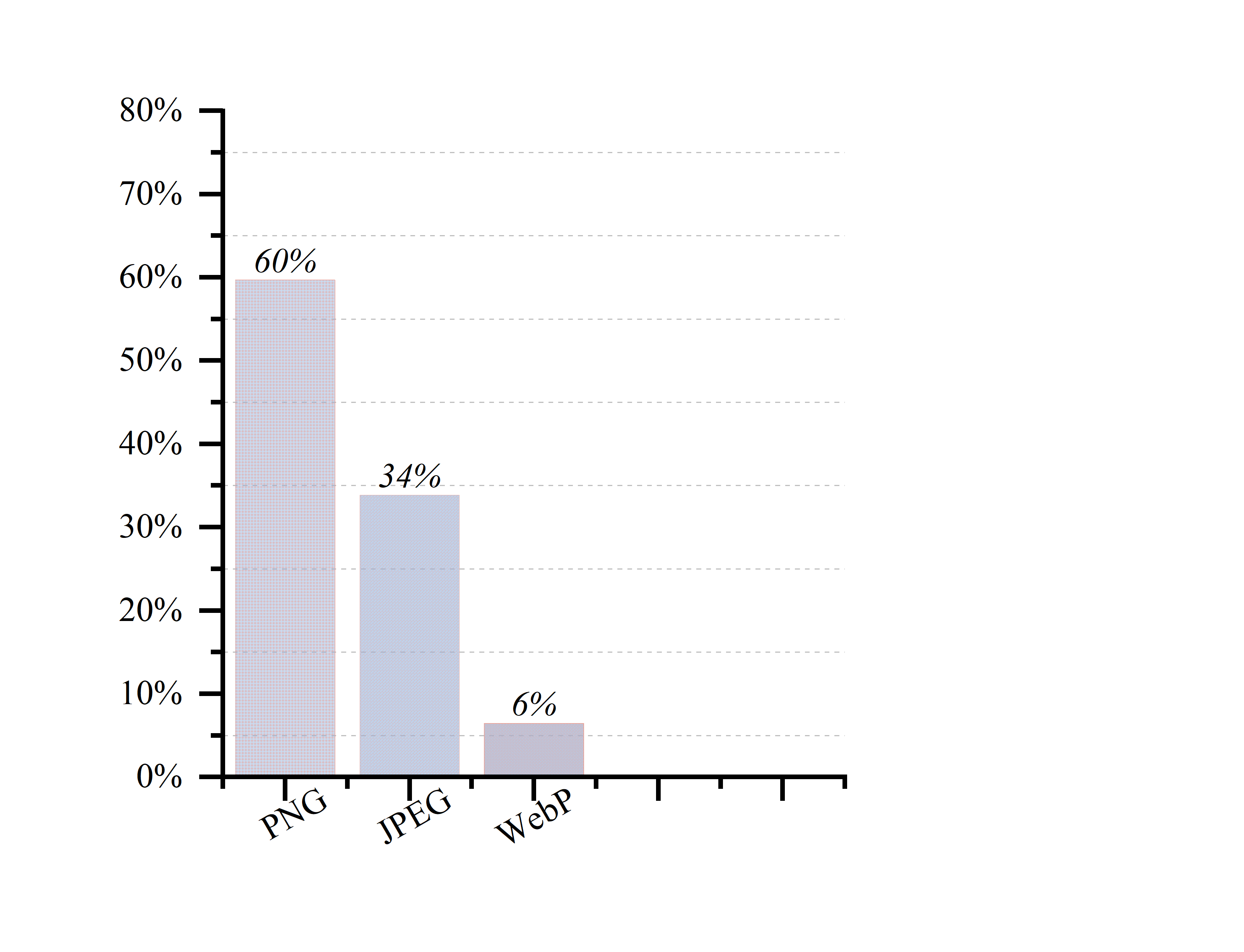} 
\caption{image formats}
\label{fig:img_dist}
\end{subfigure}
\caption{Breakdown of specific file formats for documents and images.}
\label{fig:detailed_file_types}
\end{figure}

Figure \ref{fig:detailed_file_types} illustrates the specific file format breakdown for the document and image categories.

\section{Distribution of Web Pages in DR$^{3}$-Eval's Sandbox Corpus}
\label{sec:webpage_distribution}

\begin{figure*}[htbp]
\centering

\includegraphics[width=0.8\textwidth]{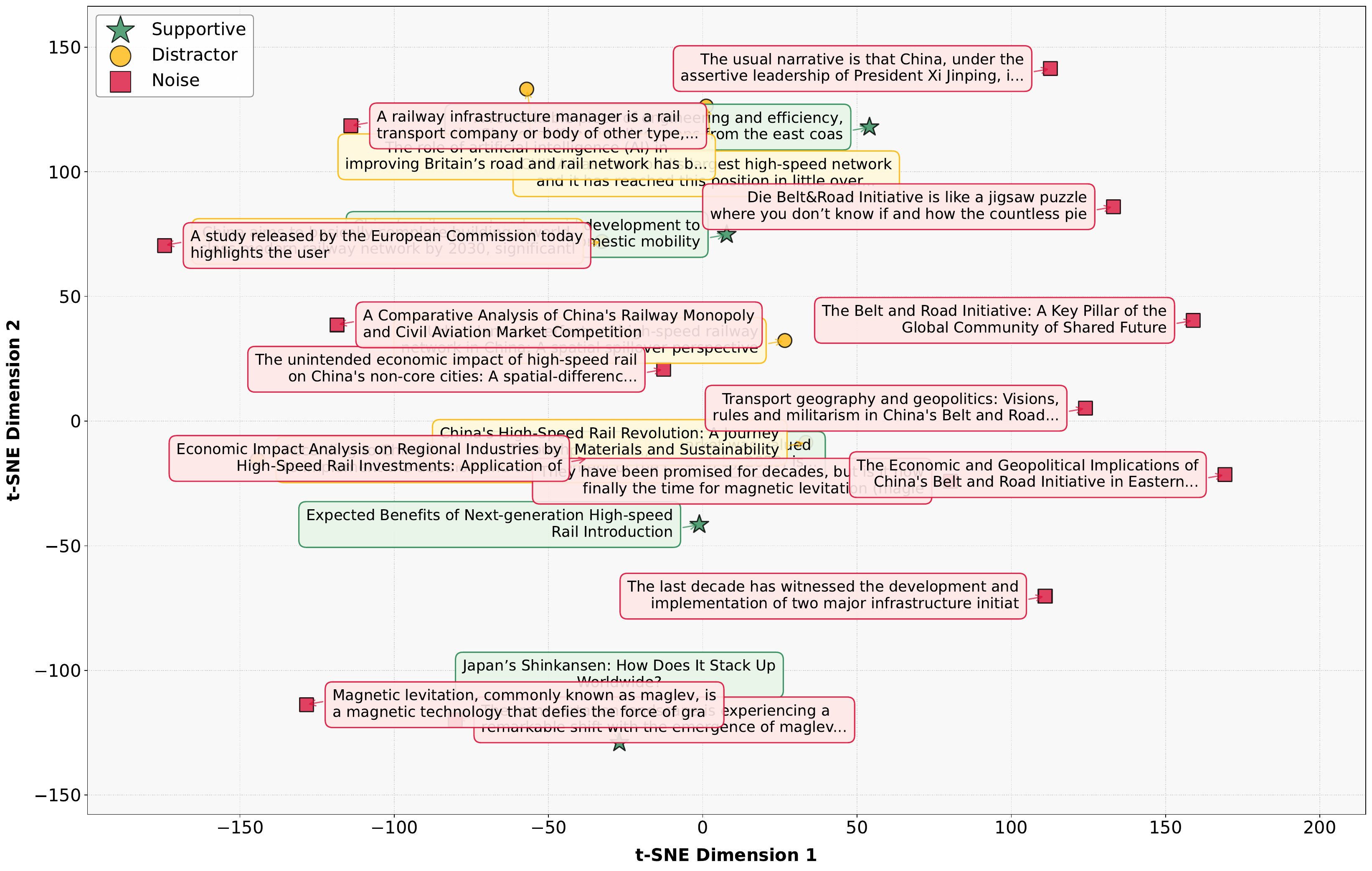}
\caption{t-SNE visualization of the semantic distribution in the Sandbox Corpus. }
\label{fig_sandbox_corpus}
\end{figure*}

To visualize the semantic distribution of the sandbox corpus, we encoded the query of a representative task and the web pages from the 64k dataset using OpenAI's \texttt{text-embedding-3-large} model. These embeddings were projected into a two-dimensional space via t-SNE. As illustrated in Figure \ref{fig_sandbox_corpus}, we plot the query relative to the documents, displaying a subset of the distractor and noise web pages to ensure visual clarity.

% From the figure \ref{fig_sandbox_corpus}, it can be seen that the semantic space distribution of various types of documents after dimensionality reduction is uniform, indicating that the constructed noise and distractor are very close to the semantic support, which can fully reflect their interference effect.

\section{Model Rankings across Different Judge LLMs}
\label{app:ranking_analysis}

Table~\ref{tab:ranking_llm_judge_appendix} compares the leaderboard rankings produced by GPT-5.1 against alternative judges (Gemini-2.5-Pro and Qwen-Max) to quantify the consistency of our evaluator selection.

\begin{table}[ht]
\centering
\caption{Comparison of rankings assigned by three different judge models. The ``Disagreement'' row indicates the number of rank swaps relative to GPT-5.1 .}
\label{tab:ranking_llm_judge_appendix}
\begin{tabular}{@{}l ccc@{}}
\toprule
\textbf{Model} & \textbf{GPT-5} & \textbf{Gemini-2.5-Pro} & \textbf{Qwen-Max} \\
\midrule
GLM-4.7         & 1 & 1 & 1 \\
Claude Sonnet 4  & 2 & 3 & 2 \\
Gemini-2.5-Pro   & 3 & 2 & 3 \\
GPT-4.1         & 4 & 5 & 5 \\
Qwen3-235B-A22B & 5 & 4 & 4 \\
Qwen3-30B-A3B  & 6 & 6 & 6 \\
\midrule
\textbf{Disagreement ($\Delta$)}  & -- & 1 & 1 \\
\bottomrule
\end{tabular}
\end{table}

\section{Human Evaluation}
\label{sec:human_evaluation}
We select 30 samples to undergo both automated and human evaluation. The human score for each sample is derived by averaging the ratings of two independent experts. The correlation coefficients between the automated and human scores are calculated as follows:
\begin{itemize}
    \item \textbf{Pearson's correlation coefficient.}
    \begin{equation*}
    r = \frac{\sum_{i=1}^{n}(A_i-A)(B_i-B)}{\sqrt{\sum_{i=1}^{n}(A_i-A)^2}\sqrt{\sum_{i=1}^{n}(B_i-B)^2}} 
    \end{equation*}
    \item \textbf{Spearman's correlation coefficient.}
    \begin{equation*}
    s = \frac{\sum_{i=1}^{n}(R(A_i)-R_A)(R(B_i)-R_B)}{\sqrt{\sum_{i=1}^{n}(R(A_i)-R_A)^2}\sqrt{\sum_{i=1}^{n}(R(B_i)-R_B)^2}} 
    \end{equation*}
    \item \textbf{Pairwise agreement.}
    \begin{equation*}
    PAR = \frac{1}{\binom{n}{2}}\sum_{i=1}^{n-1}\sum_{j=i+1}^n\mathbb{I}_{ij}
    \end{equation*}
    \begin{equation*}
        \mathbb{I}_{ij} = \begin{cases} 1 & \text{if } (A_i - A_j) \cdot (B_i - B_j) > 0 \\ 0 & \text{otherwise} \end{cases}
    \end{equation*}
\end{itemize}
where $A_i$,$B_i$ represent the automated total score and human total score for a single case, respectively.

Table \ref{tab:judge_scores} presents the fine-grained scores of the first five cases across various dimensions.

\begin{table}[h]
\centering
\caption{Comparison of LLM-Judge and Human-Judge Scores in each dimension across the first five Case}
\label{tab:judge_scores}
\begin{adjustbox}{width=0.8\linewidth, center}
\begin{tabular}{lccccccccccc}
\toprule
\multirow{2}{*}{\textbf{Case ID}} & \multicolumn{5}{c}{LLM-Judge Score} & \multicolumn{5}{c}{Human-Judge Score} \\
\cmidrule(lr){2-6} \cmidrule(lr){7-11}
 & IR$_{llm}$ & FA$_{llm}$ & DQ$_{llm}$ & IF$_{llm}$ & CC$_{llm}$ & IR$_{human}$ & FA$_{human}$ & DQ$_{human}$ & IF$_{human}$ & CC$_{human}$\\
\midrule
001 & 30.00 & 46.23 & 70.00 & 93.33 & 66.67 & 32.50 & 42.76 & 72.00 & 92.42 & 66.67 \\
002 & 40.27 & 50.38 & 70.00 & 86.67 & 66.67 & 38.68 & 53.89 & 68.00 & 83.25 & 66.67\\
003 & 42.85 & 52.25 & 70.00 & 64.00 & 16.67 & 40.78 & 56.65 & 66.86 & 62.38 & 16.67 \\
004 & 30.00 & 43.84 & 60.00 & 77.78 & 50.00 & 31.50 & 46.29 & 58.76 & 75.94 & 50.00\\
005 & 49.29 & 60.33 & 70.00 & 84.00 & 71.43 & 46.73 & 57.83 & 73.56 & 82.37 & 71.43\\
\bottomrule
\end{tabular}
\end{adjustbox}
\end{table}

\section{Report Example}
\label{sec:report_example}
We present a representative task and the corresponding report generated by DR$^{3}$-Agent. 

\begin{figure*}[h]
\centering
\includegraphics[width=\textwidth]{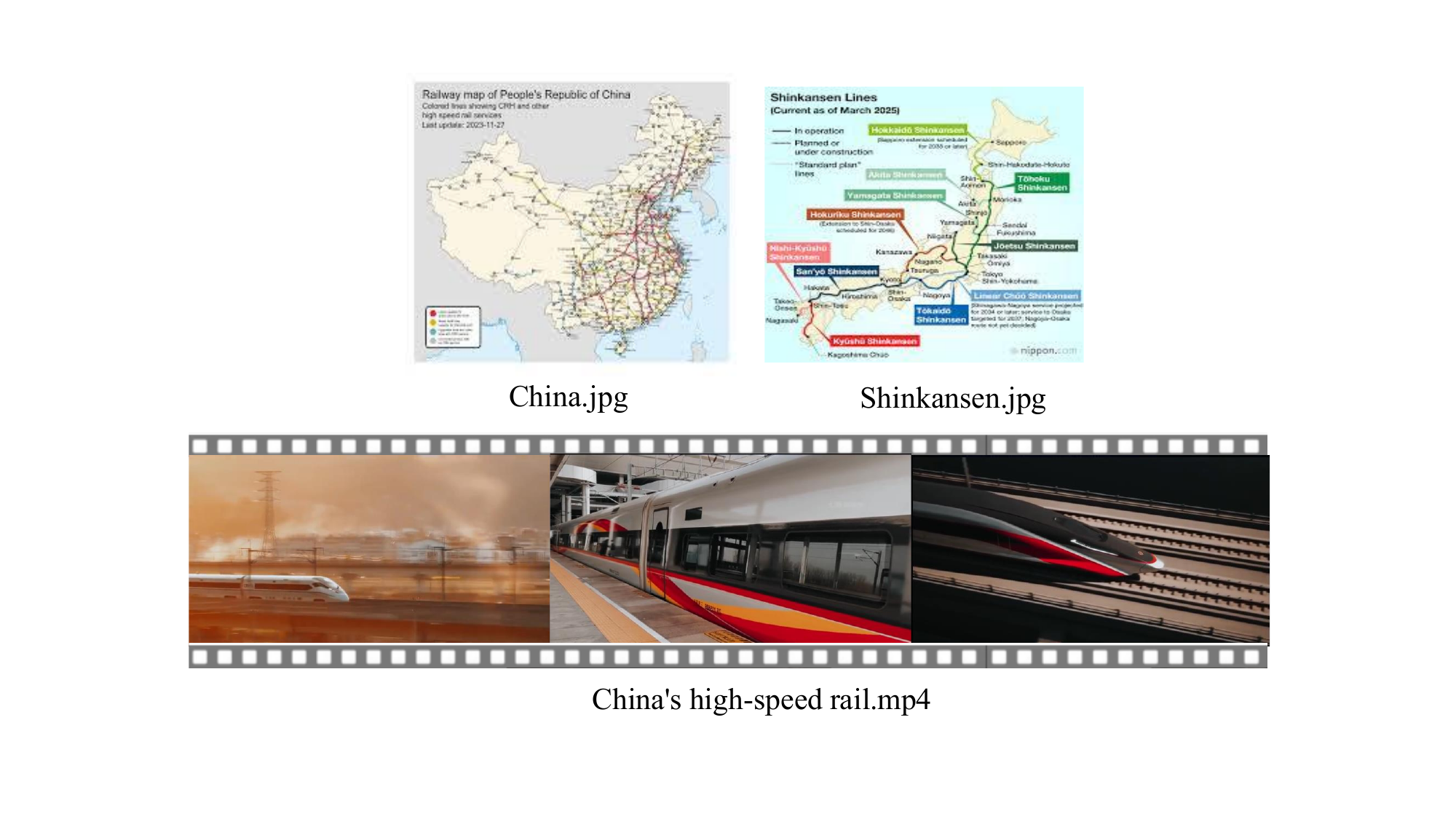}
\caption{The view of user files.}
\label{fig:user_file_view}
\end{figure*}

\textbf{User Files:}
\begin{itemize}
   \item \textbf{China.jpg} - China's high-speed rail network map showing extensive coverage and connectivity
    \item \textbf{Shinkansen.jpg} - Japan's Shinkansen network map displaying route structure and operational lines
    \item \textbf{China's high-speed rail.mp4} - Analysis of China's high-speed rail development narrative and technical achievements
\end{itemize}
Figure~\ref{fig:user_file_view} displays the visual components of the user files provided in this case study

\textbf{Query}: I've attached a video about China's high-speed rail, along with maps of China's and Japan's networks. It's impressive how fast and extensive China's system has become. Could you summarize the story behind its rapid development? I'm curious about its evolution, how its core system compares to Japan's Shinkansen, the planning behind its vast network, and what the next big steps in high-speed rail technology look like globally.

\textbf{Claude Sonnet 4's Report}
\begin{examplebox}

{\large\textbf{The Remarkable Story of China's High-Speed Rail Revolution: From Zero to Global Leader}}
\vspace{\baselineskip} 
\textbf{Executive Summary}

China's high-speed rail development represents one of the most extraordinary infrastructure transformations in modern history. In just over three decades, China evolved from having no high-speed rail capability to operating the world's largest network, fundamentally reshaping both domestic transportation and global rail technology markets. This comprehensive analysis examines the evolution of China's high-speed rail system, compares it with Japan's pioneering Shinkansen network, explores the strategic planning behind China's vast infrastructure, and identifies the next generation of rail technologies that will define the future of high-speed transportation.
\vspace{\baselineskip} 
\textbf{The Evolution: A Three-Decade Journey from Concept to Global Dominance}
\begin{itemize}
    \item \textbf{Early Foundations and Strategic Planning (1990-2007)} China's high-speed rail journey began with visionary planning in the early 1990s. In December 1990, the Ministry of Railways submitted the first proposal to build a high-speed railway between Beijing and Shanghai to the National People's Congress, marking the first time high-speed rail was proposed as an option for intercity transportation in China [``Transportation Planning Casebook/China High-Speed'']. This initial proposal set in motion a carefully orchestrated development strategy that would unfold over the following decades.
    
    The planning phase was methodical and comprehensive. In 1995, Premier Li Peng announced that preparatory work on the Beijing Shanghai HSR would begin in the 9th Five Year Plan (1996–2000), demonstrating the government's commitment to long-term infrastructure development [``Transportation Planning Casebook/China High-Speed'']. The State Council commissioned a feasibility study for the Beijing-Shanghai line in December 1994, establishing the technical and economic foundation for what would become the backbone of China's high-speed rail network [``Transportation Planning Casebook/China High-Speed''].
    
    The first operational breakthrough came in 1998 when the Guangzhou-Shenzhen line was electrified and, by introducing Swedish-made X 2000 trains, the service speed increased to 200 km/h [``Transportation Planning Casebook/China High-Speed'']. This project served as a crucial testing ground for high-speed rail operations in the Chinese context. On October 12, 2003, the Qinshen Railroad started operation as the first passenger-only railroad line in China, featuring the ``China Star'' electrified locomotive with a designed speed of 200 km/h and a top speed of 300 km/h [``Transportation Planning Casebook/China High-Speed''].
    
    \item \textbf{Strategic Technology Acquisition Phase (2004-2008)} Rather than attempting to develop high-speed rail technology entirely from scratch, China pursued a sophisticated strategy of technology acquisition and transfer. In June 2004, the Ministry of Railways accepted bids from Alstom of France, Bombardier Transportation of Germany, and Kawasaki of Japan for 200 high-speed train sets with top speed of 200 km/h, crucially including terms of technology transfer [``Transportation Planning Casebook/China High-Speed'']. This marked the beginning of systematic technology acquisition that would accelerate China's development timeline by decades.
    
    The technology transfer relationships were comprehensive and strategic. CSR obtained Japanese high-speed technology starting in 2004 as part of a deal with Kawasaki, with the CRH2 series being directly derived from the E2 Series 1000 Shinkansen and nicknamed ``Hairtail'' [``Hexie (train)'']. From June to September 2005, the Ministry of Railways purchased train sets with a top speed of 350 km/h, demonstrating the rapid escalation in technical capabilities [``Transportation Planning Casebook/China High-Speed''].
    
    A pivotal decision came in 2006 when the State Council, in its Mid-to-Long Term Railway Development Plan, adopted conventional track HSR technology over maglev, enabling rapid construction of standard gauge, passenger dedicated HSR lines throughout China [``Transportation Planning Casebook/China High-Speed'']. By April 2007, the high-speed rail system had upgraded to more than 423 km of tracks capable of supporting 250 km/h train service and 3,002 km of tracks capable of supporting 200 km/h as a result of the Speed Up Campaign [``Transportation Planning Casebook/China High-Speed''].

    \item \textbf{The Breakthrough Era (2008-2020)} The transformative moment arrived on August 1, 2008, when the Beijing-Tianjin Inter-city Railway was launched at a top speed of 350 kilometers per hour, marking the beginning of China's high-speed railway era [Video: China's high-speed rail.mp4]. This railway served as China's first independently built high-speed railway operating at 350 kilometers per hour [``On Aug. 1, 2008, the Beijing-Tianjin Inter-city Ra'']. The video analysis reveals that this achievement represented rapid development accomplished in just over 10 years, establishing China as a world leader in high-speed rail technology [Video: China's high-speed rail.mp4].
    
    The scale and pace of subsequent expansion was unprecedented in infrastructure history. When China's Premier Wen Jiabao boarded the first bullet train to open the service of the new high-speed line between Beijing and Shanghai on June 30, 2011, he described the step as a 'new chapter' in China's railway history [``When China's Premier Wen Jiabao boarded the fi'']. The 1,318km-long high-speed line connected China's two major metropolitan hubs in only four hours and 48 minutes, compared to about ten hours on the old railway [``When China's Premier Wen Jiabao boarded the fi''].
    
    The passenger adoption was remarkable. By the end of 2017, the total number of passengers who traveled on China's high-speed trains had reached 7 billion [``On Aug. 1, 2008, the Beijing-Tianjin Inter-city Ra'']. By 2020, China's total operation mileage of high-speed railway reached 30,000 kilometers, covering 80 percent of its major cities [``On Aug. 1, 2008, the Beijing-Tianjin Inter-city Ra'']. The World Bank described this as ``the biggest rail expansion the world has ever seen'' with China being ``the only country that has ever attempted such an increase'' [``When China's Premier Wen Jiabao boarded the fi''].

\end{itemize}

\textbf{Comparative Analysis: China's System vs. Japan's Shinkansen}

The comparison between China's high-speed rail network and Japan's Shinkansen reveals dramatic differences in scale, approach, and strategic objectives. The total operating mileage of China's high-speed railway network exceeds 25,000 kilometers, accounting for two thirds of the world's total [``On Aug.1, 2008, the Beijing-Tianjin Inter-city Ra'']. The China high-speed rail network map demonstrates extensive coverage that stretches from the Bohai Sea in the east to the Gobi Desert in the west, from the central plains to the mountains in the southwest, and from the snowfields in the north to the riverside towns in the south [Image: China.jpg].

The network analysis reveals dense connectivity in eastern and central regions, with key routes connecting major cities such as Beijing, Shanghai, Guangzhou, and Wuhan, forming an integrated system linking provincial capitals and major urban centers [Image: China.jpg]. The map shows a well-developed network with multiple intersecting lines and branch routes, demonstrating extensive coverage and connectivity as of November 27, 2009, with continued expansion since then [Image: China.jpg].

In contrast, Japan's Shinkansen network, while pioneering and highly efficient, covers approximately 2,800 km as of March 2025 [Image: Shinkansen.jpg]. The Japanese network shows a highly focused approach along Japan's main urban corridor, connecting major cities like Tokyo, Osaka, Nagoya, and Fukuoka through nine distinct lines: Tokaido, Sanyo, Kyushu, Tohoku, Joetsu, Hokuriku, Hokkaido, Akita, and Yamagata Shinkansen [Image: Shinkansen.jpg]. The coverage pattern is linear, following Japan's geographic constraints and population distribution, with most lines concentrated along the Pacific coast and major population centers [Image: Shinkansen.jpg].
\vspace{\baselineskip} 
\textbf{Strategic Planning Behind China's Vast Network}

China's network planning represents a revolution in infrastructure development methodology, incorporating cutting-edge technologies and comprehensive strategic thinking. Modern railway planning involves strategic oversight of all railway systems and related infrastructure, including track alignment, rail capacity, logistics planning, maintenance scheduling, safety checks, signalling and communication networks, track access, landscaping, and vegetation control [``What is railway planning? A brief overview'']. This comprehensive approach requires synchronizing human resources such as engineers, architects, urban planners, rail construction managers, transport officers, railway contractors and crews, as well as coordinating traffic, passengers, and pedestrian flow [``What is railway planning? A brief overview''].
Leaders of the best railway infrastructure worldwide are increasingly turning towards 3D virtual replicas or digital twins as a planning tool [``What is railway planning? A brief overview'']. Digital twin railway planning software generates intelligent, dynamic models of the rail network, creating a virtual representation of complex 3D environments, accurately mimicking the appearance and behaviour of real-world infrastructure [``What is railway planning? A brief overview'']. This technological approach addresses traditional inefficiencies, as an international survey by BearingPoint revealed that 81\% of rail infrastructure managers felt traditional decision-making processes were too time-consuming, 89\% believed these processes led to unnecessary duplication of effort, and 86\% were overwhelmed by paperwork duties [``What is railway planning? A brief overview''].

\vspace{\baselineskip} 
\textbf{Conclusion}

China's high-speed rail development represents a remarkable transformation that redefined global transportation infrastructure within a single generation. The journey from initial proposals in 1990 to achieving world leadership status demonstrates the power of strategic planning, systematic technology acquisition, massive investment, and unwavering political commitment. The evolution followed a clear pattern: early planning and feasibility studies (1990-2003), strategic technology acquisition through foreign partnerships (2004-2008), rapid domestic expansion (2008-2020), and emergence as a global technology exporter (2010s-present).

The comparison with Japan's Shinkansen reveals two different but successful approaches to high-speed rail development. Japan pioneered the technology with careful engineering and operational efficiency, creating a reliable and profitable system that serves as a model for precision and safety. China leveraged this foundation to create an unprecedented scale of infrastructure that prioritizes national integration and social benefits alongside transportation efficiency.

The strategic planning behind China's vast network demonstrates sophisticated integration of digital technologies, comprehensive infrastructure coordination, and long-term economic development objectives. The network serves not merely as transportation infrastructure but as a tool for national integration, regional development, and international influence.
Looking toward the future, the next generation of high-speed rail technology promises even more dramatic advances. Maglev technology offers the potential for speeds exceeding 600 km/h, while digital integration through IoT, AI, and cloud computing will revolutionize operational efficiency and passenger experience. China's continued investment in ultra-high speed conventional rail, maglev systems, and emerging technologies like hyperloop positions the country to maintain its leadership in this critical infrastructure sector.
The global expansion of high-speed rail technology, driven by Chinese companies competing with established European and Japanese manufacturers, suggests that the next decade will see accelerated adoption worldwide. As the video demonstrates with China's CR450 prototype achieving 450 km/h with improved smoothness, the pursuit of ``amazing China speed'' continues to push the boundaries of what is possible in rail transportation, setting the stage for a new era of global mobility.

\vspace{\baselineskip} 
\textbf{References}

\textbf{Video Sources:}
\begin{itemize}
    \item China's high-speed rail.mp4 - Analysis of China's high-speed rail development narrative and technical achievements
\end{itemize}

\textbf{Image Sources:}
\begin{itemize}
    \item China.jpg - China's high-speed rail network map showing extensive coverage and connectivity
    \item Shinkansen.jpg - Japan's Shinkansen network map displaying route structure and operational lines
\end{itemize}

\textbf{Database Sources:}

\begin{itemize}
    \item Transportation Planning Casebook/China High-Speed Rail development timeline and milestones
    \item Historical development of Beijing-Tianjin Inter-city Railway and network expansion
    \item Technology transfer relationships between China and international partners
    \item Railway planning methodologies and digital twin technologies
    \item Maglev technology development and global implementation
    \item Global high-speed rail market analysis and future projections
    \item Digital railways and IoT integration in transportation systems
    \item International expansion projects and technology export initiatives
\end{itemize}

\end{examplebox}

\section{Evaluation Example}
\label{sec:evaluation_example}
% This section provides a detailed introduction to the calculation process of evaluation metrics. Firstly, it presents the insights extracted from user files and sandbox corpus, and determines whether they are covered. Then, it displays the remaining metrics.
This section details the calculation of evaluation metrics, beginning with the assessment of insight coverage derived from user files and the sandbox corpus, followed by the computation of the remaining metrics.

\subsection{Information Recall from User Files and Sandbox Corpus}
\subsubsection{Information Recall from User Files}
\textbf{Golden insight from user files}
\begin{enumerate}
    \item China's high-speed rail network is dense, especially in the east.
    \item The map shows the extensive network as of November 2023.
    \item The network includes lines with speeds of 300 km/h or more.
    \item Rail lines are color-coded by speed, from $<200$ to $\ge 300$~km/h.
    \item Map of Japan's Shinkansen lines as of March 2025.
    \item Shows operational, planned, and under-construction routes.
    \item A future Linear Chūō Shinkansen (maglev) line is projected.
    \item The network connects major cities like Tokyo, Osaka, and Hakata.
    \item Developed from non-existent to world-class in just over 10 years.
    \item Current trains travel at world-leading speeds of 300-350 km/h.
    \item The new CR450 EMU prototype is the world's fastest.
    \item CR450 prototype reaches 450 km/h in tests.
\end{enumerate}

\begin{table}[h]
\centering
\caption{Evaluation of Information Recall from User Files.}
\label{tab:user-files}
% 调整行间距，让表格更舒展
\renewcommand{\arraystretch}{1.5} 
\begin{tabular}{
    >{\centering\arraybackslash}m{0.8cm}   % 第一列：水平+垂直居中
    >{\centering\arraybackslash}m{3cm}     % 第二列：水平+垂直居中
    >{\raggedright\arraybackslash}m{11cm}  % 第三列：左对齐+垂直居中
}
\toprule
\textbf{Number} & \textbf{Status} & \textbf{Evidence} \\ 
\midrule
1 & \textcolor{blue}{Covered} & The network analysis reveals dense connectivity in eastern and central regions, with key routes connecting major cities... \\ 
\hline
2 & \textcolor{gray}{Half Covered} &  The map shows a well-developed network ... as of November 27, 2009, with continued expansion since then [Image: China.jpg].\\
\hline
3 & \textcolor{blue}{Covered} &  The Beijing-Tianjin Inter-city Railway was launched at a top speed of 350 kilometers per hour.'' ``CRH380A reaching up to 380 km/h...\\
\hline
4 & \textcolor{red}{Not Covered} &  Not Found\\
\hline
5 & \textcolor{blue}{Covered} & Japan's Shinkansen network ... covers approximately 2,800 km as of March 2025 [Image: Shinkansen.jpg]. \\
\hline
6 & \textcolor{gray}{Half Covered} & Japan's Shinkansen network ... connecting major cities ... through nine distinct lines... \\
\hline
7 & \textcolor{gray}{Half Covered} & The experimental Japanese L0 Series maglev set a train speed record of 603 km/h (375 mph) in 2015... \\
\hline
8 & \textcolor{blue}{Covered} & connecting major cities like Tokyo, Osaka, Nagoya, and Fukuoka through nine distinct lines...\\
\hline
9 & \textcolor{blue}{Covered} & The video analysis reveals that this achievement represented rapid development accomplished in just over 10 years, establishing China as a world leader in high-speed rail technology. \\
\hline
10 & \textcolor{gray}{Half Covered} & Beijing-Tianjin Inter-city Railway ... top speed of 350 kilometers per hour.'' ``CRH380A reaching up to 380 km/h.'' ``Shinkansen ... up to 320 km/h... \\
\hline
11 & \textcolor{gray}{Half Covered} & the CR450 EMU prototype that reaches 450 km/h in testing, representing a significant advancement in speed capabilities\\
\hline
12 & \textcolor{blue}{Covered} & the CR450 EMU prototype that reaches 450 km/h in testing\\
\bottomrule
\end{tabular}
\end{table}

\subsubsection{Information Recall from Sandbox Corpus}

\textbf{Extracted insight from sandbox corpus:}
\begin{enumerate}
    \item Reducing aerodynamic resistance is crucial for faster trains.
    \item Shinkansen's strengths are efficiency and passenger comfort.
    \item China has an ambitious 2035 high-speed rail expansion plan.
    \item Digital transformation is key to future rail network evolution.
    \item Future rail relies on IoT, 5G, and AI technologies.
    \item China plans to extend its HSR network to Southeast Asia.
\end{enumerate}

\begin{table}[htbp]
\centering
\caption{The Calculation of Information Recall from Sandbox Corpus.}
\label{tab:user-files}
% 调整行间距，让表格更舒展
\renewcommand{\arraystretch}{1.5} 
\begin{tabular}{
    >{\centering\arraybackslash}m{0.8cm}   % 第一列：水平居中 + 垂直居中
    >{\centering\arraybackslash}m{3cm}     % 第二列：水平居中 + 垂直居中
    >{\raggedright\arraybackslash}m{11cm}  % 第三列：左对齐 + 垂直居中
}
\toprule
\textbf{Number} & \textbf{Status} & \textbf{Evidence} \\ 
\midrule
1 & \textcolor{red}{Not Covered} &  Not Found\\
\hline
2 & \textcolor{gray}{Half Covered} & Japan... has established a reliable and profitable system, serving as a paradigm of precision and safety. \\
\hline
3 & \textcolor{red}{Not Covered} & The medium and long-term railway development plan from 2004 to 2020... plans to invest...'' and so on, only up to the investment planning for 2025. \\
\hline
4 & \textcolor{blue}{Covered} &  Digital twins... as planning tools... address the inefficiencies of traditional decision-making... Digital railways encompass real-time data collection, predictive modeling, automated decision-making, and more.\\
\hline
5 & \textcolor{blue}{Covered} & Leading this evolution are various IoT-related technologies, such as 5G, AI, and cloud computing... which play a pivotal role in digital logistics and transportation. \\
\hline
6 & \textcolor{gray}{Half Covered} & China's high-speed rail technology is going global... projects such as the Ankara-Istanbul high-speed rail, the Jakarta-Bandung high-speed rail, and the Moscow-Kazan high-speed rail. \\
\bottomrule
\end{tabular}
\end{table}

\subsection{Citation Coverage}

\begin{longtable}{
    >{\centering\arraybackslash}m{0.8cm}  % 序号列：居中，固定宽度
    >{\raggedright\arraybackslash}m{0.62\linewidth} % 标题/内容列：左对齐，适配行宽
    >{\centering\arraybackslash}m{2.5cm}  % 状态列：居中，固定宽度
}
\caption{Evaluation of Citation Coverage.} 
\label{tab:citation-coverage}\\
\toprule
\textbf{No.} & \textbf{Source Title} & \textbf{Status} \\
\midrule
\endfirsthead

% --- 续页表头 ---
\multicolumn{3}{@{}l}{\tablename~\thetable\ (Continued)} \\
\toprule
\textbf{No.} & \textbf{Source Title} & \textbf{Status} \\
\midrule
\endhead

% --- 页脚设置 ---
\midrule
\multicolumn{3}{r@{}}{\small (Continued on next page)} \\
\endfoot

\bottomrule
\endlastfoot

% --- 第一部分：Web Page 来源（标注分类，与User File区分）---
\multicolumn{3}{l}{\textit{\textbf{Web Page Coverage}}} \\
\midrule
1 & Japan’s Shinkansen: How Does It Stack Up Worldwide? & \textcolor{blue}{\textbf{Cited}} \\ 
\midrule
2 & The global rail transportation market was valued at US\$ 724,180 million in 2022 and, by 2029, is pro & \textcolor{blue}{\textbf{Cited}} \\ 
\midrule
3 & Expected Benefits of Next-generation High-speed Rail Introduction. & \textcolor{red}{Missed} \\
\midrule
4 & China’s railway network: rapid development to support domestic mobility & \textcolor{red}{Missed} \\
\midrule
5 & A remarkable piece of engineering and efficiency, the Shanghai-Kunming line spans from the east coas & \textcolor{red}{Missed} \\
\midrule

% --- 第二部分：User File 来源（新增3行，独立分类标注）---
\multicolumn{3}{l}{\textit{\textbf{User File Coverage}}} \\
\midrule
6 & China.jpg & \textcolor{blue}{\textbf{Cited}} \\
\midrule
7 & Shinkansen.jpg & \textcolor{blue}{\textbf{Cited}} \\
\midrule
8 & China's high-speed rail.mp4 & \textcolor{blue}{\textbf{Cited}} \\

\end{longtable}

\subsection{Factual Accuracy}
% 确保导言区有这些包
% \usepackage{longtable}
% \usepackage{booktabs}
% \usepackage{array}
% \usepackage{pifont} % 用于 \cmark 和 \xmark
% \newcommand{\cmark}{\ding{51}}
% \newcommand{\xmark}{\ding{55}}

% 确保导言区有这些包
% \usepackage{longtable}
% \usepackage{booktabs}
% \usepackage{array}
% \usepackage{pifont} % 用于 \cmark 和 \xmark
% \newcommand{\cmark}{\ding{51}}
% \newcommand{\xmark}{\ding{55}}

% 确保导言区有这些包
% \usepackage{longtable}
% \usepackage{booktabs}
% \usepackage{array}
% \usepackage{pifont} % 用于 \cmark 和 \xmark
% \newcommand{\cmark}{\ding{51}}
% \newcommand{\xmark}{\ding{55}}

\begin{longtable}{
    >{\centering\arraybackslash}m{0.8cm}  % 序号列
    >{\raggedright\arraybackslash}m{13cm} % 文本列
    >{\centering\arraybackslash}m{1.5cm}  % 状态列
}
\caption{The Evaluation of Factual Accuracy.}
\label{tab:checklist}\\
\toprule
\textbf{No.} & \textbf{Claim to be verified} & \textbf{Status} \\
\midrule
\endfirsthead

% --- 续页表头 ---
\multicolumn{3}{@{}l}{\tablename~\thetable\ (Continued)} \\
\toprule
\textbf{No.} & \textbf{Claim to be verified} & \textbf{Status} \\
\midrule
\endhead

% --- 页脚设置 ---
\midrule
\multicolumn{3}{r@{}}{\small (Continued on next page)} \\
\endfoot

\bottomrule
\endlastfoot

% --- 表格内容 ---
% 1. 修复 railwa -> railway ...
1 & In December 1990, the Ministry of Railways submitted the first proposal to build a high-speed railway... & \cmark \\ 
\midrule
% 2. 添加 ...
2 & In 1995, Premier Li Peng announced that preparatory work on the Beijing Shanghai HSR would begin in... & \cmark \\
\midrule
% 3. 添加 ...
3 & The State Council commissioned a feasibility study for the Beijing-Shanghai line in December 1994, e... & \cmark \\
\midrule
% 4. 添加 ...
4 & The first operational breakthrough came in 1998 when the Guangzhou-Shenzhen line was electrified and... & \cmark \\
\midrule
% 5. 修复 lin -> line ...
5 & On October 12, 2003, the Qinshen Railroad started operation as the first passenger-only railroad line... & \cmark \\
\midrule
% 6. 修复 Transportatio -> Transportation ...
6 & In June 2004, the Ministry of Railways accepted bids from Alstom of France, Bombardier Transportation... & \cmark \\
\midrule
% 7. 添加 ...
7 & CSR obtained Japanese high-speed technology starting in 2004 as part of a deal with Kawasaki, with t... & \cmark \\
\midrule
% 8. 添加 ...
8 & From June to September 2005, the Ministry of Railways purchased train sets with a top speed of 350 k... & \cmark \\
\midrule
% --- 中间省略部分 ---
\multicolumn{3}{c}{$\vdots$} \\
\midrule
% 48. 添加 ...
48 & Planned investments in 2025 alone total 590 billion yuan (approximately \$80.8 billion) to develop an... & \xmark \\
\midrule
% 49. 修复 province -> provinces ... (推测)
49 & The expansion of China's rail network benefits regions in the interior of the country, with provinces... & \xmark \\
\midrule
% --- 后续省略部分 ---
\multicolumn{3}{c}{$\vdots$} \\
\midrule
% 72. 添加 ...
72 & As the video demonstrates with China's CR450 prototype achieving 450 km/h with improved smoothness,... & \cmark \\

\end{longtable}

\subsection{Format Compliance}

\begin{longtable}{
    >{\centering\arraybackslash}m{0.8cm}  % 新增：序号列，居中固定宽度，不挤压内容
    >{\raggedright\arraybackslash}m{12.8cm}% 调整：要求列，拉宽适配页面，保留左对齐
    >{\centering\arraybackslash}m{1.5cm}   % 保留：状态列，居中放置符号，宽度不变
}
\caption{The Evaluation of Format Compliance.}
\label{tab:checklist_compliance}\\
\toprule
\textbf{No.} & \textbf{Requirement} & \textbf{Satisfied} \\  % 表头新增序号列，格式统一
\midrule
\endfirsthead

% --- 续页表头（同步新增序号列）---
\multicolumn{3}{@{}l}{\tablename~\thetable\ (Continued)} \\
\toprule
\textbf{No.} & \textbf{Requirement} & \textbf{Satisfied} \\
\midrule
\endhead

% --- 页脚设置（列数同步调整为3）---
\midrule
\multicolumn{3}{r@{}}{\small (Continued on next page)} \\
\endfoot

\bottomrule
\endlastfoot

% --- 表格内容（新增连续序号，文本完全未改动）---
1 & Mention China's high-speed rail rapid development & \cmark \\ 
\midrule
2 & Describe early stages of China's high-speed rail & \cmark \\
\midrule
3 & Describe key phases in China's network expansion & \cmark \\
\midrule
4 & Explain main drivers of China's rapid rail growth & \cmark \\
\midrule
5 & Mention role of national planning in China & \cmark \\
\midrule
6 & Describe overall layout of China's high-speed network & \cmark \\
\midrule
7 & Explain planning logic behind China's network structure & \cmark \\
\midrule
8 & Describe core technical system of China's high-speed rail & \cmark \\
\midrule
9 & Describe core technical system of Japan's Shinkansen & \cmark \\
\midrule
10 & Compare technical features of China rail and Shinkansen & \cmark \\
\midrule
11 & Compare operational characteristics of China rail and Shinkansen & \cmark \\
\midrule
12 & Explain key differences between China rail and Shinkansen & \cmark \\
\midrule
13 & Mention future global high-speed rail technology trends & \cmark \\
\midrule
14 & Describe next big technological steps in high-speed rail & \cmark \\
\midrule
15 & Explain global implications of upcoming high-speed rail advances & \cmark \\

\end{longtable}
\subsection{Depth Quality}
\noindent\textbf{Score:} 0.70

\vspace{0.5em} % 增加一点间距

\noindent\textbf{Justification:} 
The report gives a solid, well-structured narrative of China's HSR evolution. It clearly explains the technology-transfer strategy, compares China's system to Japan's on scale, speed, and philosophy, and touches on planning tools and future technologies like maglev and digital railways.

However, much of the ``planning'' section drifts into generic global railway-planning and IoT discussion rather than concrete Chinese institutional planning choices (e.g., corridor selection, financing models, governance reforms, safety scandals). Furthermore, the comparison with Shinkansen lacks deeper critical analysis of safety records, cost overruns, demand risk, and long-term economic performance.

The future-tech section is descriptive and somewhat market-report-like, with limited critical assessment of feasibility, trade-offs, and timelines. Consequently, the analysis is adjudged as good but not yet truly deep or nuanced.

\section{Inference and Evaluation Cost}
\label{cost}

We report the approximate runtime and API cost of the full pipeline.

\begin{table}[h]
\centering
\small
\begin{tabular}{lcc}
\toprule
\textbf{Setting} & \textbf{Avg. time / task} & \textbf{Approx. API cost / task} \\
\midrule
Inference  & 300--400 s & \$0.3--1.0\$ \\
Evaluation & 90 s        & \$0.10\$ \\
\bottomrule
\end{tabular}
\caption{Approximate runtime and API cost of the full pipeline.}
\label{tab:cost}
\end{table}
\section{Prompts for Data Construction}
\label{app:data_prompts}
This section outlines the prompts used to synthesize our dataset.

\subsection{Search Terms Generation}
\label{search terms generation}
\begin{tcolorbox}[
    title=Prompt for Search Terms Generation, % 标题
    colback=gray!5,    % 背景色：极浅灰
    colframe=black!75, % 边框色：深灰
    breakable          % 允许跨页
]
 You are a professional research assistant. Based on the following user files, generate search terms for web search.

\textbf{User Files}

{files\_summary}

\textbf{Task Requirements}

\textbf{Generate 10 Search Terms}

Based on the content of the user files, generate 10 \textbf{English} search terms for web search. These search terms should:
\begin{itemize}
    \item Be able to retrieve supplementary information related to the topic of the user files
    \item Cover different angles and aspects (such as background knowledge, case studies, policies and regulations, technical details, industry trends, etc.)
    \item \textbf{Each search term should be 3-5 words} - combining 2-3 related concepts naturally
    \item \textbf{Use common, widely-searched phrases} that are likely to return 100+ search results
    \item Search terms should be able to find \textbf{information not in the user files but helpful for understanding the topic}
    \item  \textbf{All 10 terms must be ORTHOGONAL} - they should target completely different aspects and NOT overlap in search results
\end{itemize}

\textbf{CRITICAL RULES for generating search terms that return many results:}
\begin{enumerate}
    \item  \textbf{Balance specificity with searchability}
    \begin{itemize}
        \item  If the topic is about a specific country/region (e.g., Japan tax policy), include it
        \item If the topic is general, prefer broader terms that apply globally
        \item BAD: ``Japan income tax bracket structure 2024" (too specific + year)
        \item GOOD:``Japan income tax reform" or ``income tax policy reform"
    \end{itemize}
    \item \textbf{Use organization names strategically}
    \begin{itemize}
        \item Include organization names when they are central to the topic
        \item Avoid obscure acronyms that few people search for
        \item BAD: ``HUD FMR calculation methodology" (obscure acronym)
        \item GOOD: ``HUD housing affordability" or ``federal housing assistance"
    \end{itemize}
    \item \textbf{Prefer common terms over academic jargon}
    \begin{itemize}
        \item Use terms that practitioners and general public would search
        \item BAD: ``arthrofibrosis prophylaxis protocols" (too technical)
        \item GOOD: ``joint stiffness prevention therapy"
    \end{itemize}
    \item \textbf{Use common search patterns}:
    \begin{itemize}
        \item ``[Topic] + [Aspect]": ``smart agriculture technology"
        \item ``[Topic] + [Method]": ``precision farming data analytics"
        \item ``[Topic] + [Trend]": ``renewable energy market growth"
        \item ``[Topic] + [Impact]": ``climate change economic effects"
        \item ``[Topic] + [Impact]": ``climate change economic effects"
    \end{itemize}
\end{enumerate}

\textbf{Good Examples (3-5 words, will return 100+ results):}
\begin{itemize}
    \item ``smart agriculture technology development"
    \item ``precision farming data analytics"
    \item ``vaccine efficacy clinical research"
    \item ``housing affordability policy reform"
    \item ``pension system sustainability analysis"
    \item ``renewable energy investment trends"
    \item ``supply chain risk management"
    \item ``Japan pension reform challenges" (specific country when relevant)
    \item ``OECD education policy recommendations" (organization when relevant)
\end{itemize}

\textbf{Bad Examples (too specific or obscure):}
\begin{itemize}
    \item ``Japan corporate tax rate competitiveness analysis 2024" (too long, includes year)
    \item ``Fair Market Rent determination methodology HUD" (obscure + too specific)
    \item ``California DMV licensing procedures step by step" (too specific)
    \item ``pension actuarial valuation discount rate assumptions" (too technical)
\end{itemize}

\textbf{Language Requirements}

\textbf{Important}: All output content (search\_terms, topic\_summary) must be in \textbf{English}.

\textbf{Output Format}

Please output in JSON format:

```json

{{

  "search\_terms": [
  
    "search term 1",
    
    "search term 2",
    
    "search term 3",
    
    "search term 4",
    
    "search term 5",
    
    "search term 6",
    
    "search term 7",
    
    "search term 8",
    
    "search term 9",
    
    "search term 10"
    
  ],
  
  "topic\_summary": "Brief summary of the user files topic (1-2 sentences)"
  
}}
```
Only output JSON, no other content.
\end{tcolorbox}

\subsection{Query Construction}
\label{query construction}

\begin{tcolorbox}[
    title=Prompt for Query Generation, % 标题
    colback=gray!5,    % 背景色：极浅灰
    colframe=black!75, % 边框色：深灰
    breakable          % 允许跨页
]
\textbf{Task}
   
Generate a natural user query based on user files, search results, and keyword info.

Core Goal: The generated query should naturally guide an AI agent to search for "relevant keywords" without directly exposing them.

\textbf{User Files}

{chr(10).join(f'- {name}' for name in user\_file\_names)}

\textbf{Relevant Keywords (what agent should search)}

{relevant\_kw\_text}

\textbf{Noise Keywords (what agent should NOT be guided to search)}

{noise\_kw\_text}

\textbf{Keyword Classification Reasoning}

{reasoning}

\textbf{Search Results ({len(useful\_search)} total)}

{search\_text}

\textbf{Query Design Requirements}

\textbf{Core Principles}
\begin{enumerate}
    \item \textbf{Concise}: Query must be SHORT (50-100 words), like a real user's brief question, not verbose
    \item \textbf{Natural}: Query should be from user's perspective, like a real person would ask
    \item \textbf{Guiding}: Query topic should naturally lead agent to search ``relevant keywords", but don't over-hint
    \item  \textbf{No Exposure}: Don't directly use technical terms from keywords, use simple natural expressions
    \item \textbf{Brief File Reference}: Query must briefly mention user files, like ``based on my xxx file" or ``see attachment"
    \item \textbf{Cover All Results}: Query must be designed so ALL {len(useful\_search)} search results are needed for a complete answer, even if each result is only used a little
    \item \textbf{Use All Files}: Query must be designed so ALL {len(user\_file\_names)} user files are needed for a complete answer, even if each file is only used a little
\end{enumerate}

\textbf{Design Approach}
\begin{enumerate}
    \item Analyze the common theme of relevant keywords
    \item Design a SHORT natural query (50-100 words), don't over-describe background
    \item Query should:
    \begin{itemize}
        \item Be short and direct, like a casual question
        \item Not contain technical jargon or hint-like words
        \item Briefly mention user files
    \end{itemize}
\end{enumerate}

\textbf{Examples}
If relevant keywords are:
- ``Three-distance method spatial layout modern pocket park design cases"
- ``Scattered perspective step-by-step scenery urban micro-renewal"

User file is:

- Suzhou\_Garden\_Design.pdf

\xmark BAD Query (too long, over-hinting):

``Based on my Suzhou Garden Design.pdf showing spatial layout techniques, help me research how these traditional garden concepts are applied in contemporary urban small green spaces, please find some recent practice cases"

\xmark BAD Query (directly exposes keywords):

``Please analyze the application cases of three-distance method in modern pocket parks"

\cmark GOOD Query (short, natural):

``Based on my Suzhou garden design file, help me find some similar style applications in modern small parks"

\cmark GOOD Query (short, natural):

``With the attached garden diagram, research how traditional garden techniques are practiced in urban small green spaces"

\textbf{Output Format}

Output JSON only:\\
{{ \\
  ``query": ``Complete user query (50-100 words, short and natural, mentions user files)",\\
  ``file\_references": [``User files referenced in query and their content description"],\\
  ``subtasks": [\\
    {{\\
      ``id": 1,\\
      ``description": ``Subtask description",\\
      ``target\_keywords": [``Relevant keywords this subtask should guide agent to search"]\\
    }}\\
  ],\\
  ``design\_rationale":``Why this query guides agent to search relevant keywords",\\
  ``avoided\_terms": [``Technical terms deliberately avoided in query"],\\
  ``natural\_expressions": [``Natural expressions used to replace technical terms"]\\
}}\\

Note: Output JSON only, no other content. Query must include clear reference to user files!

\end{tcolorbox}

\section{Prompts for Evaluation Preparation}

\subsection{Insights Extraction from User Files}
\label{Insights Extraction from User Files}
\begin{tcolorbox}[
    title=Prompt for Insights Extraction from User Files, % 标题
    colback=gray!5,    % 背景色：极浅灰
    colframe=black!75, % 边框色：深灰
    breakable          % 允许跨页
]
 Please extract \textbf{Key Insights} relevant to the query from the following web search results.

\textbf{Query Task}

{query}

\textbf{Important Notes (Web Search Results)}

You are processing web search results. These sources contain key information needed to answer the query. Please extract \textbf{the core information provided by each source}.

\textbf{Core Principle}

\textbf{Source Specificity}: Extract information that is \textbf{unique and specific to each web page}, NOT common knowledge or generic statements.

\textbf{Extraction Requirements}
\begin{enumerate}
    \item \textbf{Source Specificity (MOST IMPORTANT)}:
    \begin{itemize}
        \item  \cmark Extract \textbf{specific information unique to this source}, such as: specific names, dates, data, events, product names, company names, project names, method names, etc.
        \item \cmark Extract \textbf{unique viewpoints, conclusions, methods, or findings} from this source
        \item \xmark \textbf{DO NOT} extract common knowledge, such as: ``AI can improve efficiency", ``data analysis is important", ``user experience matters"
        \item \xmark \textbf{DO NOT} extract generic descriptions, such as: ``this method works well", ``performance improved", ``solved the problem"
    \end{itemize}
    \item \textbf{Source Contribution}: Extract the main contribution of each source to answering the query, such as:
    \begin{itemize}
        \item  Methods/techniques/concepts introduced by the source (e.g., ``proposes AfME embedding", ``uses MCMC optimization")
        \item Core topics or problems discussed by the source
        \item  Key conclusions or findings of the source
        \item \textbf{Note}: No need to extract precise numbers (e.g., ``19.86\% improvement"), just extract core concepts (e.g., ``significant PSNR improvement")
    \end{itemize}
    \item \textbf{Relevance}: Information is relevant to the query task
    \item \textbf{Verifiability}: Can determine whether the report mentions this information (semantic similarity is sufficient, exact match not required)
    \item \textbf{Atomicity}: Each insight must be atomic, containing only 1-12 words, expressing a simple fact or concept
\end{enumerate}

\textbf{Examples of Common Knowledge (DO NOT Extract)}
\begin{itemize}
    \item ``Machine learning requires large amounts of data" → This is common knowledge
    \item ``User experience is important" → This is common knowledge
    \item ``This method improved accuracy" → Too vague, no specific value or comparison
    \item ``Further research is needed" → This is a generic conclusion
\end{itemize}

\textbf{Examples of Source-Specific Content (SHOULD Extract)}
\begin{itemize}
    \item ``Uses ResNet-50 as backbone network" → Specific technical choice
    \item ``Achieves 92.3\% accuracy on ImageNet" → Specific data
    \item ``Proposed by John Smith in January 2024" → Specific name and time
    \item ``Employs AfME embedding method" → Specific method name
\end{itemize}

\textbf{Quantity Control}
\begin{itemize}
    \item \textbf{Extract 1-2 core insights per source (web page)}
    \item Ensure each source has corresponding insights
    \item Prioritize conceptual information over precise numbers
    \item If a source lacks specific information, extract fewer insights rather than extracting common knowledge
\end{itemize}

\textbf{Language Requirement}

\textbf{Important}: Output insights must be in English.

\textbf{Output Format}

Please output insights in JSON array format, each insight should include the source title:

```json

[

  {{``insight": ``Brief source-specific information point (1-12 words)", ``source": ``Source Title"}}
  
]

```

Only output the JSON array, no other content.

{content\_text}

\end{tcolorbox}

\subsection{Insights Extraction from Sandbox Corpus}
\label{Insights Extraction from Sandbox Corpus}
\begin{tcolorbox}[
    title=Prompt for Insights Extraction from Sandbox Corpus, % 标题
    colback=gray!5,    % 背景色：极浅灰
    colframe=black!75, % 边框色：深灰
    breakable          % 允许跨页
]
 Please extract \textbf{Key Insights} relevant to the query from the following web search results.

\textbf{Query Task}

{query}

\textbf{Important Notes (Web Search Results)}

You are processing web search results. These sources contain key information needed to answer the query. Please extract \textbf{the core information provided by each source}.

\textbf{Core Principle}

\textbf{Source Specificity}: Extract information that is \textbf{unique and specific to each web page}, NOT common knowledge or generic statements.

\textbf{Extraction Requirements}
\begin{enumerate}
    \item \textbf{Source Specificity (MOST IMPORTANT)}:
    \begin{itemize}
        \item  \cmark Extract \textbf{specific information unique to this source}, such as: specific names, dates, data, events, product names, company names, project names, method names, etc.
        \item \cmark Extract \textbf{unique viewpoints, conclusions, methods, or findings} from this source
        \item \xmark \textbf{DO NOT} extract common knowledge, such as: ``AI can improve efficiency", ``data analysis is important", ``user experience matters"
        \item \xmark \textbf{DO NOT} extract generic descriptions, such as: ``this method works well", ``performance improved", ``solved the problem"
    \end{itemize}
    \item \textbf{Source Contribution}: Extract the main contribution of each source to answering the query, such as:
    \begin{itemize}
        \item Methods/techniques/concepts introduced by the source (e.g., ``proposes AfME embedding", ``uses MCMC optimization")
        \item Core topics or problems discussed by the source
        \item Key conclusions or findings of the source
        \item \textbf{Note}: No need to extract precise numbers (e.g., ``19.86\% improvement"), just extract core concepts (e.g., ``significant PSNR improvement")
    \end{itemize}
    \item \textbf{Relevance}: Information is relevant to the query task
    \item \textbf{Verifiability}: Can determine whether the report mentions this information (semantic similarity is sufficient, exact match not required)
    \item \textbf{Atomicity}: Each insight must be atomic, containing only 1-12 words, expressing a simple fact or concept
\end{enumerate}

\textbf{Examples of Common Knowledge (DO NOT Extract)}
\begin{itemize}
    \item ``Machine learning requires large amounts of data" → This is common knowledge
    \item ``User experience is important" → This is common knowledge
    \item ``This method improved accuracy" → Too vague, no specific value or comparison
    \item ``Further research is needed" → This is a generic conclusion
\end{itemize}

\textbf{Examples of Source-Specific Content (SHOULD Extract)}
\begin{itemize}
    \item ``Uses ResNet-50 as backbone network" → Specific technical choice
    \item ``Achieves 92.3\% accuracy on ImageNet" → Specific data
    \item ``Proposed by John Smith in January 2024" → Specific name and time
    \item ``Employs AfME embedding method" → Specific method name
\end{itemize}

\textbf{Quantity Control}
\begin{itemize}
    \item \textbf{Extract 1-2 core insights per source (web page)}
    \item Ensure each source has corresponding insights
    \item Prioritize conceptual information over precise numbers
    \item If a source lacks specific information, extract fewer insights rather than extracting common knowledge
\end{itemize}

\textbf{Language Requirement}

\textbf{Important}: Output insights must be in English.

\textbf{Output Format}

Please output insights in JSON array format, each insight should include the source title:

```json

[
  {{``insight": ``Brief source-specific information point (1-12 words)",
  
  ``source": ``Source Title"}}
  
]

```

Only output the JSON array, no other content.

{content\_text}

\end{tcolorbox}

\subsection{Checklist Generation}
\label{Checklist Generation}

% \begin{checklistpromptbox}
\begin{tcolorbox}[
    title=Prompt for Checklist Generation, % 标题
    colback=gray!5,    % 背景色：极浅灰
    colframe=black!75, % 边框色：深灰
    % fontupper=\ttfamily, % 内容字体：小号打字机字体
    breakable          % 允许跨页
]
    Extract atomic evaluation requirements from the following query. Each requirement must be short, specific, and independently verifiable.

Query:

{query\_text}

Rules:
\begin{enumerate}
    \item \textbf{Atomic Decomposition}: Break down complex requirements into minimal, independent checkpoints
    \begin{itemize}
        \item Each requirement checks only one specific point
        \item Example: ``Analyze aspects A, B, and C" → Split into ``Mention A", ``Mention B", ``Mention C"
        \item Example: ``Compare X and Y" → Split into ``Describe X", ``Describe Y", ``Explain differences"
    \end{itemize}
    \item \textbf{Short and Clear}: Each requirement should be one sentence, no more than 10 words
    \begin{itemize}
        \item Bad: ``Based on the uploaded innovation theory article and e-commerce comparison chart, analyze Temu's business model"
        \item Good: ``Mention innovation theory", ``Cite e-commerce data", ``Analyze Temu model"
    \end{itemize}
    \item \textbf{Independently Verifiable}: Each requirement can be judged individually (yes/no)
    \begin{itemize}
        \item Bad: ``Deeply analyze multiple dimensions of the business model"
        \item Good: ``Mention value proposition", ``Mention revenue sources", ``Mention cost structure"
    \end{itemize}
    \item \textbf{Cover Implicit Requirements}: Extract implicit checkpoints from the query
    \begin{itemize}
        \item If ``compare" is mentioned → Need ``Describe A", ``Describe B", ``Explain differences"
        \item If ``determine whether" is mentioned → Need ``Provide evidence", ``Draw conclusion"
        \item If ``refer to article" is mentioned → Need ``Cite article content"
    \end{itemize}
    \item  \textbf{Quantity Target}: A query should typically have 8-15 atomic requirements
    \begin{itemize}
        \item Simple query: 8-10 items
        \item Medium query: 10-12 items
        \item Complex query: 12-15 items
        \item Keep moderate granularity, not too few or too many
    \end{itemize}
    \item \textbf{Categories}:
    \begin{itemize}
        \item content: Mention a specific content point
        \item evidence: Cite a material/data source
        \item analysis: Perform some analysis
        \item comparison: Make a comparison
        \item conclusion: Draw a conclusion
    \end{itemize}
    \item \textbf{Format Requirements}:
    \begin{itemize}
        \item Start with verbs: ``Mention", ``Cite", ``Analyze", ``Compare", ``Explain", ``Describe"
        \item Avoid compound sentences, one thing per item
        \item Each item no more than 10 words
    \end{itemize}
\end{enumerate}

Output JSON:

{{

    ``checklist": [
    
        {{``id": 1, ``requirement": ``Mention X", ``category": ``content"}},
        
        {{``id": 2, ``requirement": ``Cite Y data", ``category": ``evidence"}},
        
        {{``id": 3, ``requirement": ``Analyze Z features", ``category": ``analysis"}},
        
        ...
        
    ]
    
}}
\end{tcolorbox}

\section{Prompts for Evaluation }
\label{evaluation prompt}

We utilize an LLM-as-a-Judge approach to score the model's responses

\subsubsection{Information Recall}
\label{sec:information_recall_prompt} 
\begin{tcolorbox}[
    title=Prompt for Evaluation of Information Recall, % 标题
    colback=gray!5,    % 背景色：极浅灰
    colframe=black!75, % 边框色：深灰
    breakable          % 允许跨页
]
You are an intelligent assistant that labels atomic insights based on whether they can be extracted from a given report.

\textbf{Search Query/Task}

{self.query or ``No specific query provided"}

\textbf{Insights to Label (Total: {len(batch)})}

{insights\_list}

\textbf{Report/Passage}

{result\_text}

TASK: Label each insight as 1.0 (support), 0.5 (Partial Support), or 0.0 (Not Support) based on whether the insight can be extracted from the report.

TERNARY SCORING SYSTEM 

[SCORE 1.0 - FULLY COVERED]

Definition: The report EXPLICITLY and COMPLETELY covers the insight's core meaning.

Requirements for 1.0 (ALL must be met):
\begin{itemize}
    \item \cmark The core information point is clearly stated in the report
    \item \cmark No significant details are missing
    \item \cmark The meaning is unambiguous and complete
    \item \cmark A reader would understand the full insight from the report alone
\end{itemize}

Examples of 1.0:
\begin{itemize}
        \item Insight: ``In 2023, Shanghai achieved a waste classification coverage rate of 95\%"
        
      Report: ``The coverage rate of household waste classification in Shanghai in 2023 has reached over 95\%" → 1.0 (exact match)
    \item  Insight: ``EPNP algorithm for camera pose estimation"
    
  Report: ``Using EPNP algorithm for camera pose estimation" → 1.0 (same meaning, different wording)
\end{itemize}

[SCORE 0.5 - PARTIALLY COVERED]

Definition: The report contains RELATED content but is INCOMPLETE or VAGUE.

Conditions for 0.5 (any of these):
\begin{enumerate}
    \item Only part of the core meaning is covered (missing key details)
    \item The topic is mentioned but specifics are absent
    \item Related concept exists but not the exact point
    \item Generalization without the specific insight
    \item The connection requires inference (not explicit)
\end{enumerate}

Examples of 0.5:
\begin{itemize}
    \item Insight: ``Shanghai's garbage classification coverage rate will reach 95\% by 2023"
    
  Report: ``Shanghai's garbage classification has achieved significant results" → 0.5 (topic covered, but no specific percentage)
    \item Insight: ``Germany adopts a dual track recycling system"
    
  Report: ``Developed countries have mature recycling systems" → 0.5 (generalization, no specific country/system)
    \item Insight: ``AI ethics need to establish accountability mechanisms"
    
  Report: ``The development of AI needs to consider ethical issues" → 0.5 (related topic, but no specific mechanism)
\end{itemize}

[SCORE 0.0 - NOT COVERED]

Definition: The report does NOT contain the insight's information.

Conditions for 0.0:
\begin{itemize}
    \item \xmark The topic is completely absent from the report
    \item \xmark Only unrelated keywords appear
    \item \xmark The information contradicts the insight
    \item \xmark No semantic connection exists
\end{itemize}

Examples of 0.0:
\begin{itemize}
    \item Insight: ``Singapore's waste incineration power generation technology"
    
  Report: (never mentions Singapore) → 0.0
    \item Insight: ``The specific content of the 666 training plan"
    
  Report: (never mentions ``666" or this training plan) → 0.0
\end{itemize}
THREE-STEP JUDGMENT PROCESS

For EACH insight, follow these steps:

\textbf{STEP 1: Extract Core Meaning}
\begin{itemize}
    \item Identify the KEY information points (usually 1-3)
    \item Note any specific details (numbers, names, dates, methods)
    \item  Example: ``Shanghai's garbage classification coverage rate will reach 95\% by 2023" 
  → Core: [Shanghai, 2023, garbage classification, coverage rate, 95\%]
\end{itemize}

\textbf{STEP 2: Search Report for Coverage}
\begin{itemize}
    \item Look for exact matches first
    \item Then look for semantic equivalents
    \item Note what IS found and what is MISSING
\end{itemize}

\textbf{STEP 3: Apply Scoring Rules}
\begin{itemize}
    \item If ALL core points are explicitly covered → 1.0
    \item If SOME core points are covered OR topic is mentioned without specifics → 0.5
    \item If NO core points are found → 0.0
\end{itemize}

SEMANTIC EQUIVALENCE GUIDELINES 

\cmark The following are considered SEMANTICALLY EQUIVALENT (score 1.0):

a) Synonym Substitution:
\begin{itemize}
    \item ``Law" $\approx$ ``Regulations" $\approx$ ``Regulations" $\approx$ ``Provisions"
    \item ``Significant results" $\approx$ ``Significant effects" $\approx$ ``Success achieved"
    \item ``Diversified technology" $\approx$ ``Multiple technologies" $\approx$``Rich variety of technologies"
\end{itemize}

b) Hypernym/Hyponym:
\begin{itemize}
    \item ``EPNP algorithm" → ``PnP algorithm"(EPNP is a type of PnP) → 0.5 (too general)
    \item ``Germany, Japan" → ``Developed Countries ``/" International Experience" → 0.5 (generalization)
    \item ``Gaode Map" → ``Navigation Application"(if context is clear) → 0.5 (less specific)
\end{itemize}

c) Paraphrasing :
\begin{itemize}
    \item ··Establish AI Ethics Review Committee" $\approx$ ``Establish AI Ethics Committee for Review" → 1.0
    \item ··Six ethical design details" $\approx$ ``Six major ethical issues" → 1.0 (if all 6 are covered)
\end{itemize}

d) Information Fusion:
\begin{itemize}
    \item Insight content is integrated into a larger discussion
    \item Core information points are preserved with different expression
\end{itemize}

\xmark The following are NOT semantically equivalent (score 0.0 or 0.5):

a) Completely Different Topics:
\begin{itemize}
    \item Insight discusses topic A, report discusses topic B
    \item No semantic overlap whatsoever
\end{itemize}

b) Keyword Overlap Only:
\begin{itemize}
    \item Same words appear but with completely different meanings
    \item Example: Insight says ``Apple company", report says ``apple fruit"
\end{itemize}

c) Missing Information:
\begin{itemize}
    \item Core information points from insight are completely absent in report
    \item  Example: Insight mentions specific number ``666", report never mentions it
\end{itemize}

d) Contradictory Information:
\begin{itemize}
    \item Report information contradicts the insight
    \item Example: Insight says ``successful", report says ``failed"
\end{itemize}

BOUNDARY CASE HANDLING 

When uncertain, follow this priority:

\begin{enumerate}
    \item If >50\% of core meaning is covered → 1.0
    \item If reasonable semantic connection exists → 1.0
    \item If only weak connection or keyword overlap → 0.5
    \item If no connection at all → 0.0
\end{enumerate}

Principle: Prefer false positives over false negatives
(The goal of recall assessment is to check if information is missing)

% DECISION BOUNDARY EXAMPLES

% | Insight | Report Content | Score | Reason |

% |---------|---------------|-------|--------|

% |Shanghai 95\% coverage rate | Shanghai 95\% coverage rate | 1.0 | Exact match|

% |Shanghai has a 95\% coverage rate | Shanghai has a very high coverage rate | 0.5 | Missing specific number|

% |95\% coverage rate in Shanghai | Progress of garbage classification in China | 0.5 | Related topic, no Shanghai detail|

% |Shanghai 95\% coverage rate | (not mentioned) | 0.0 | Absent|

% |EPNP algorithm | PnP algorithm | 0.5 | Related but not specific variant|

% |EPNP Algorithm | EPNP Algorithm | 1.0 | Exact Match|

% |German Dual Track System | International Experience | 0.5 | Generalization|

% |'Germany's Dual Track System' | 'Germany's Dual Track System Recycling' | 1.0 | Specific Match|

% |Legal provisions | Regulatory requirements | 1.0 | Synonym substitution|

% |Establish Committee | Establish Committee | 1.0 | Paraphrasing|

RESPONSE FORMAT

Respond ONLY with valid JSON (no markdown, no extra text):

{{

    ``results": [
    
        {{
        
            ``id": 1,
            
            ``core\_points": [``point1", ``point2"],
            
            ``found\_in\_report": ``[quote or describe what was found]",
            
            ``missing\_points": [``what is missing, if any"],
            
            ``score": 1.0,
            
            ``reasoning": ``[brief reason in Chinese, <30 words]"
            
        }},
        
        {{
        
            ``id": 2,
            
            ``core\_points": [``point1", ``point2"],
            
            ``found\_in\_report": ``Related content but incomplete",
            
            ``missing\_points": [``specific detail X"],
            
            ``score": 0.5,
            
            ``reasoning": ``[brief reason in Chinese, <30 words]"
            
        }},
        
        {{
        
            ``id": 3,
            
            ``core\_points": [``point1"],
            
            ``found\_in\_report": ``NOT FOUND",
            
            ``missing\_points": [``all points missing"],
            
            ``score": 0.0,
            
            ``reasoning": ``[brief reason in Chinese, <30 words]"
            
        }}
        
    ]
    
}}

You MUST provide exactly {len(batch)} results. Use ONLY scores 1.0, 0.5, or 0.0.
        
\end{tcolorbox}

\subsubsection{Factual Accuracy}
\label{sec:factual_accuracy_prompt} 
\begin{tcolorbox}[
    title=Prompt for Evaluation of Factual Accuracy , % 标题
    colback=gray!5,    % 背景色：极浅灰
    colframe=black!75, % 边框色：深灰
    breakable          % 允许跨页
]
\textbf{Task}
    
You are a fact-checking expert. Please verify whether the following statement is \textbf{relevant and not contradictory} to the content of the source document.

\textbf{Important principle: Maintain the stability of the evaluation, avoid fluctuations, and adopt a relatively lenient evaluation standard. Only consider a result as false if there are obvious errors.}

\textbf{Statements to be verified}

{claims\_text}

\textbf{Validation Criteria}

\textbf{Cases where the determination is supported: true}

As long as \textbf{any one of the following conditions} is met, it is determined to be true:
\begin{enumerate}
    \item \textbf{Any part} of the statement has relevant content in the source document
    \item A statement is a \textbf{generalization, summary, inference, or extension} of the content of the source document
    \item The statements use different wording, but \textbf{have similar semantics}
    \item The statement contains \textbf{implicit information} from the source document
    \item For images/videos: The content described may be visually visible or inferable
    \item The statement is a \textbf{reasonable interpretation} of the content of the source document, even if it is not the only interpretation
    \item The source document contains \textbf{partially supporting} content for this statement
\end{enumerate}

\textbf{Situations where it is determined as supported: }

false (limited to the following situations)

Only when one of the following conditions is met, it is determined as false:
\begin{enumerate}
    \item Statements that are \textbf{directly contradictory} to the source document (such as significant errors in numbers or completely opposite facts)
    \item The source document \textbf{completely lacks} any relevant content stated
    \item The statement cannot be reasonably inferred from the source document
\end{enumerate}

\textbf{Judgment principles}
\begin{itemize}
    \item Allow \textbf{substantial generalization and inference}
    \item Allow \textbf{wording differences} and \textbf{different ways of expression}
    \item Allow \textbf{partially correct} statements (as long as they are not completely wrong)
    \item For situations that are \textbf{ambiguous or uncertain}, they should all be determined as true
    \item For multimodal content, \textbf{subjective description} and \textbf{reasonable imagination} are allowed
\end{itemize}

\textbf{Example}
\begin{itemize}
    \item Source document: ``The company's revenue increased by 25\% in 2023"
    \begin{itemize}
        \item ``The company's revenue has seen a significant increase" → true (generalization)
        \item ``The company's performance is good" → true (inferred)
        \item ``The company's revenue in 2023 increased by approximately 25\%." → true (approximately)
        \item ``The company's revenue increased by about 55\% in 2023" → false (not true)
        \item ``The company's revenue in 2023 decreased by 25\%" → false (direct contradiction)
    \end{itemize}
    \item Source document: A product image
    \begin{itemize}
        \item ``The product's exterior design is exquisite" → true (subjective description)
        \item ``The product adopts a modern design style" → true (reasonable inference)
    \end{itemize}
\end{itemize}

\textbf{Output format (strict JSON)}

{{``results": [{{``id": 1, ``supported": true, ``explanation": ``Brief explanation"}}]}}

Please start the verification:

\end{tcolorbox}

\subsubsection{Instruction Following}
\label{sec:instruction_following_prompt} 
\begin{tcolorbox}[
    title=Prompt for Evaluation of Instruction Following, % 标题
    colback=gray!5,    % 背景色：极浅灰
    colframe=black!75, % 边框色：深灰
    breakable          % 允许跨页
]
 Check if the report satisfies each requirement. Output ONLY true/false for each item.

\textbf{Requirements ({batch\_size} items)}

{checklist\_list}

\textbf{Report}

{result\_text}

\textbf{Rules}
\begin{itemize}
    \item true: Report clearly addresses this requirement with specific content
    \item false: Not mentioned, or only briefly/superficially mentioned
\end{itemize}

\textbf{Response Format (JSON only, no explanation needed)}

{{``results": [{``, ".join([f'{{``id": {i+1}, ``satisfied": true/false}}' for i in range(batch size)])}]}}

\end{tcolorbox}

\subsubsection{Depth Quality}
\label{sec:depth_quality_prompt} 
\begin{tcolorbox}[
    title=Prompt for Evaluation of Depth Quality, % 标题
    colback=gray!5,    % 背景色：极浅灰
    colframe=black!75, % 边框色：深灰
    breakable          % 允许跨页
]
 You are an expert Research Report Evaluator.

        You are given:
        \begin{enumerate}
            \item A research report
            \item A research question that the report attempts to answer
        \end{enumerate}
        
        \texttt{\textless research\_question\textgreater}
        
        {Question}
        
       \texttt{\textless/research\_question\textgreater}

        \texttt{\textless Report\textgreater}
        
        {result\_text}
        
        \texttt{\textless/Report\textgreater}

        \textbf{Instructions:}

        \textbf{ANALYZE THOROUGHLY}: Examine the report in detail and identify any issues, even small ones. Look for subtle problems, minor inconsistencies, areas that could be improved, or any shortcomings that might affect the quality.

        Evaluate the report according to the \textbf{Depth \& Quality of Analysis} criterion. Provide:

        \begin{itemize}
            \item A \textbf{score between 1 and 10} (must be an integer) using the scale defined below.
            \item A \textbf{detailed justification} (2–3 sentences) in \textbf{simple plain English} explaining why you gave that score, including any specific issues or strengths you identified.
        \end{itemize}

        \textbf{General Scoring Scale (1-10, integers only):}
        \begin{itemize}
            \item \textbf{1} = Completely Unusable: No meaningful content, completely off-topic, or gibberish
            \item \textbf{2} = Severely Deficient: Major fundamental flaws, almost no useful information
            \item \textbf{3} = Very Poor: Significant problems throughout, barely addresses the topic
            \item \textbf{4} = Poor: Multiple major issues, lacks coherence or depth, many gaps
            \item \textbf{5} = Below Average: Notable weaknesses, incomplete coverage, superficial analysis
            \item \textbf{6} = Adequate: Meets minimum requirements but with clear limitations and room for improvement
            \item \textbf{7} = Good: Solid work that addresses the topic well, minor issues present
            \item \textbf{8} = Very Good: High quality with comprehensive coverage, few minor issues
            \item \textbf{9} = Excellent: Outstanding work with deep insights, nearly flawless execution
            \item \textbf{10} = Perfect: Exceptional, publication-ready quality with innovative insights
        \end{itemize}

        \textbf{CRITICAL EVALUATION REQUIREMENTS}:
        \begin{enumerate}
            \item \textbf{Use the FULL scoring range}: Distribute scores across 1-10 based on actual quality differences. Do NOT cluster scores in a narrow range.
            \item \textbf{Differentiate clearly}: A mediocre report should score 4-5, a good report 6-7, an excellent report 8-9. Only truly exceptional work deserves 10.
            \item \textbf{Be discriminating}: Look for specific quality differences between reports. Better analysis, clearer structure, and deeper insights should result in higher scores.
            \item \textbf{Penalize appropriately}: Minor issues = small deductions (0.5-1 point), major issues = significant deductions (2-3 points).
            \item \textbf{Reward excellence}: If a report demonstrates exceptional depth, clarity, or insight, give it the high score it deserves.
            \item \textbf{Compare mentally}: Consider how this report compares to the best and worst possible reports on this topic.
        \end{enumerate}

        \textbf{Evaluation Criterion: \textbf{Depth \& Quality of Analysis}}
        
        Evaluate how thoroughly the report analyzes the research question. \textbf{BE HARSH}: Look for superficiality, missing details, lack of evidence, weak reasoning.

        \begin{itemize}
            \item \textbf{1-2}: Completely superficial, no real analysis, just lists facts
            \item \textbf{3-4}: Very basic analysis, misses most key factors, no depth at all
            \item \textbf{5}: Basic analysis present but very incomplete, misses many important aspects
            \item \textbf{6}: Covers main points but analysis is shallow, lacks nuance and deeper insights
            \item \textbf{7}: Adequate analysis with some depth, but still missing sophistication or critical thinking
            \item \textbf{8}: Good depth with multiple factors explored, shows some sophisticated reasoning (RARE)
            \item \textbf{9}: Exceptional depth with nuanced understanding and insights (VERY RARE)
            \item \textbf{10}: Perfect analysis with groundbreaking insights and comprehensive understanding (ALMOST NEVER)
        \end{itemize}

        ------------------

        \textbf{Output format:}

        \texttt{\textless evaluation\textgreater}
        
        \texttt{\textless depth\_quality\textgreater}
        
           \texttt{\textless score\textgreater}
           
            1–10 (integer only, based on the scoring scale above)
            
            \texttt{\textless/score\textgreater}
            
            \texttt{\textless justification\textgreater}
            
            Give a detailed 2–3 sentence justification for your score in simple plain English, including specific issues or strengths.
            
            \texttt{\textless/justification\textgreater}
            
        \texttt{\textless/depth\_quality\textgreater}
        
        \texttt{\textless/evaluation\textgreater}

\end{tcolorbox}

\end{document}